\documentclass[lettersize,journal]{IEEEtran}
\usepackage[table]{xcolor}
\usepackage{amsmath,amsfonts}
\usepackage{algorithmic}
\usepackage{algorithm}
\usepackage{array}
\usepackage[caption=false,font=normalsize,labelfont=sf,textfont=sf]{subfig}
\usepackage{textcomp}
\usepackage{stfloats}
\usepackage{url}
\usepackage{verbatim}
\usepackage{graphicx}
\usepackage{cite}
\usepackage{booktabs}
\usepackage{booktabs}
\usepackage{xcolor}
\usepackage{colortbl}
\usepackage{multirow}
\usepackage{makecell}
\usepackage{bm}
\usepackage{tabularx}
\definecolor{myLightBlue}{HTML}{D6EAF8} 
\definecolor{myLightRed}{HTML}{FADBD8}

\begin{document}

\title{AD-SRF: Attack-Defense Semantic Reinforcement \\ for Multi-Modal Image Fusion}

\author{Hao Zhang, Zhiqian Zheng, Linfeng Tang, Xinyu Xiang, and Jiayi Ma, \IEEEmembership{Senior Member, IEEE }
	\thanks{This work was supported by the XXX. \textit{(Hao Zhang and Zhiqian Zheng contributed equally to this work.) (Corresponding author: Jiayi Ma.)}}

\IEEEcompsocitemizethanks{\IEEEcompsocthanksitem H. Zhang, Z. Zheng, X. Xiang, and J. Ma are with the Electronic Information School, Wuhan University, Wuhan, 430072, China; L. Tang is with the School of Robotics, Wuhan University, Wuhan, 430072, China; H. Zhang is also with the Suzhou Institute, Wuhan University, Suzhou, 215123, China, and J. Ma is also with the School of Robotics, Wuhan University, Wuhan, 430072, China (e-mail: zhpersonalbox@gmail.com, zqzheng1105@gmail.com, linfeng0419@gmail.com, xiangxinyu@whu.edu.cn, jyma2010@gmail.com).}
}

\markboth{IEEE Transactions on XXXX}%
{Zhang \MakeLowercase{\textit{et al.}}: AD-SRF: Attack-Defense Semantic Reinforcement for Multi-Modal Image Fusion}


\maketitle

\begin{abstract}
Existing image fusion methods struggle to balance visual quality and semantic attributes due to the inherent conflict between their optimization objectives. To address this challenge, we propose an attack-defense semantic reinforcement framework for multi-modal image fusion, termed AD-SRF. By decoupling the optimization of these two aspects and then losslessly reintegrating them, this framework injects rich semantic attributes into the fused image without compromising its visual quality. First, we develop a transformer-based restorative visual fusion network as a strong visual baseline. Equipped with the cross-modal retrieval module and the information restoration–fusion collaborative loss, it produces highly visible fused images even under degraded conditions. Second, we introduce an attack-defense game mechanism to strengthen the semantic attributes of the fused images. The attacker learns perturbation patterns that impair downstream segmentation, whereas the defender restores missing semantics from the original inputs and reintegrates them into the fused image through an imperceptible semantic map. Through iterative adversarial interaction, this mechanism drives the defender to continually strengthen the fused image’s semantic attributes, boosting downstream segmentation performance without compromising visual quality. Extensive experiments demonstrate that AD-SRF achieves a win-win performance in both visual quality and semantic attributes. The code is publicly available at \url{https://github.com/HaoZhang1018/AD-SRF}.
\end{abstract}

\begin{IEEEkeywords}
Image fusion, attack-defense game, semantic reinforcement, restoration.
\end{IEEEkeywords}

\section{Introduction}
\label{sec:intro}
\IEEEPARstart{G}{iven} the complexity of real-world environments, a single imaging sensor often captures only partial scene information. Such limitations hinder reliable perception from the information source, making the scene neither clearly visible nor reliably understood. In this context, the multi-modal image fusion (MMIF) technology has emerged as an effective means to address the limitation~\cite{zhang2021image,liu2024infrared}. By combining complementary information from multiple modalities, a single fused image can be generated that provides a comprehensive scene representation. Owing to these compelling advantages, MMIF technology has been extensively applied in a variety of fields, \textit{e.g.}, remote sensing monitoring, intelligent security, and autonomous driving~\cite{zhou2024probing,li2025dual,gong2023sifdrivenet}.

\begin{figure}[t]
	\centering
	\includegraphics[width=1\linewidth]{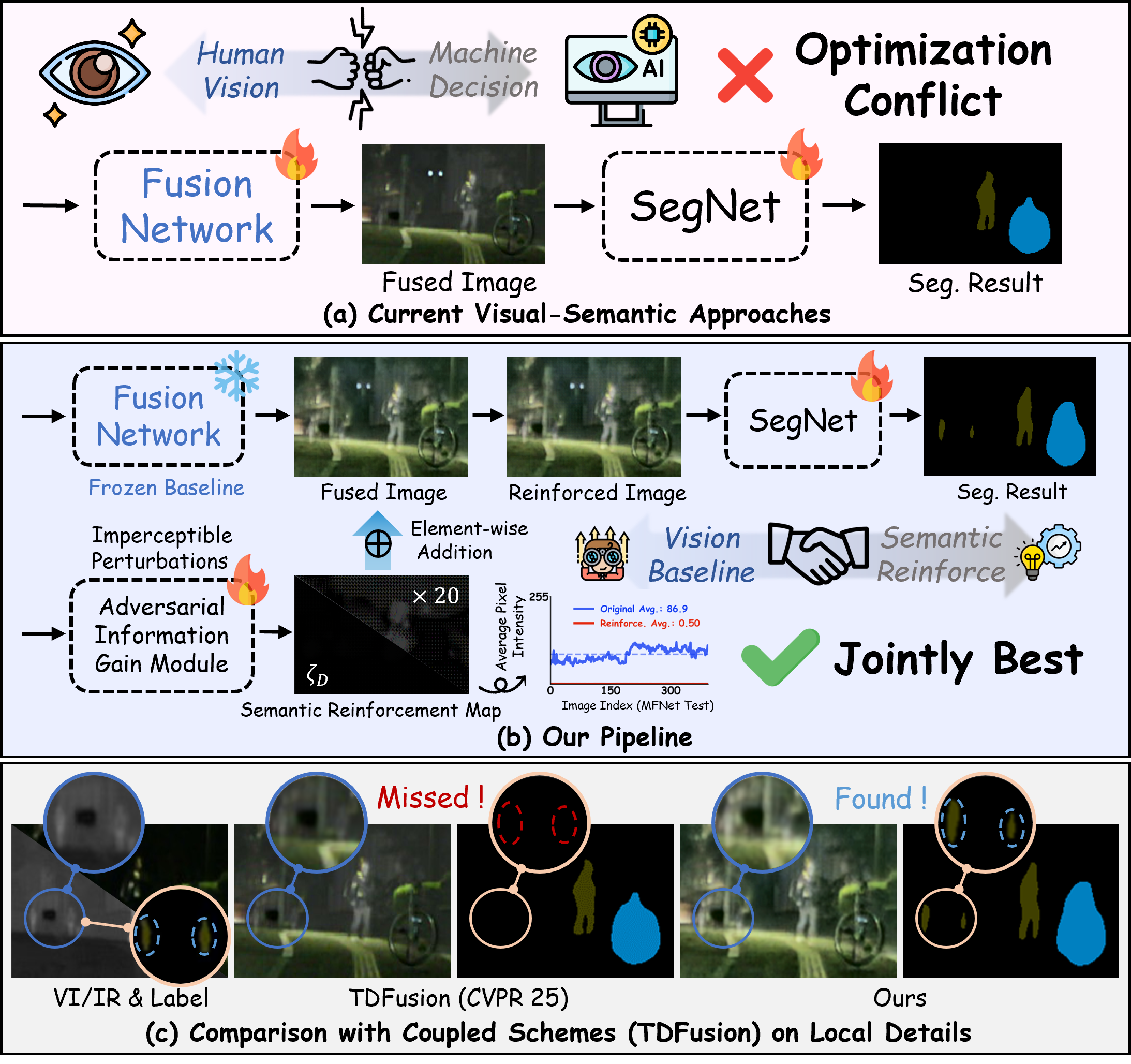}
	\vspace{-0.2in}
	\caption{Comparative analysis of existing coupled and our proposed decoupled paradigms. Our AD-SRF achieves both high-fidelity visual representations and accurate semantic decision performance under adverse conditions, whereas the SOTA competitor suffers from weakened visualization and missed detections.}\label{fig:1}
\end{figure}

When developing MMIF methods, two key aspects should be considered. (1) \textit{Visual quality}. Scene information should be effectively recovered from degradations, producing fused images with rich textures and significant contrast for high-quality human visual observation~\cite{zhang2025omnifuse,zhao2025freefusion,Zhou2025A}. (2) \textit{Semantic attributes}. Fused images should preserve rich semantic information to support accurate and reliable decision-making by downstream intelligent systems~\cite{tang2022image,liu2022target,liu2023multi}. Therefore, it is highly desirable to develop MMIF methods that achieve both high visual quality and rich semantic attributes. However, most existing methods can barely simultaneously deal with these two aspects.

\subsection{Previous Arts}
In recent years, the swift development of deep learning has provided a sustained impetus for the advancement of MMIF, giving rise to a variety of innovative methods. According to their optimization guidance, these methods can be divided into the following two categories.

\textbf{Purely Visual Driven Methods}. The majority of existing MMIF methods belong to this category, employing various types of mechanisms to ensure the visual enhancement of fused images. For instance, U2Fusion~\cite{xu2022u2fusion} introduces an adaptive intensity and gradient loss that dynamically retains texture structures and significant thermal objects, producing promising visual results. CDDFuse~\cite{zhao2023cddfuse} proposes a complementary feature decomposition strategy that dynamically adjusts the fusion weights based on inter-feature correlations, thereby achieving effective integration of multi-modal appearance characteristics. Mask-DiFuser~\cite{tang2025mask} transforms the image fusion task into a dual-mask image reconstruction problem, driving the neural network to extract high-definition features from multi-modal images. By further leveraging a diffusion model to learn natural image priors, this method ensures that the fused image aligns more closely with human visual perception. Furthermore, OmniFuse~\cite{zhang2025omnifuse} focuses on the impact of degradations on the fusion process, seamlessly integrating degradation removal and information fusion in the latent space. This enables the fused results to provide high-quality visual representations even under challenging conditions, such as low light and noise. Although the above methods can satisfy the visual observation of humans, they fail to account for the preservation of semantic attributes, making the fused images less suitable for downstream intelligent decision-making tasks, \textit{e.g.}, semantic segmentation, object detection.

\textbf{Hybrid Visual–Semantic Driven Methods}. Recently, the MMIF community has recognized the above issue and begun shifting towards the development of hybrid visual–semantic driven methods, aiming to enhance visual quality while also preserving high-quality semantic attributes. For example, SeAFusion~\cite{tang2022image} and SegMiF~\cite{liu2023multi} employ a segmentation model to guide the optimization of the fusion model, improving the semantic attributes of the fused image through the weighted and iterative updates of visual and semantic loss terms. Similarly, TarDAL~\cite{liu2022target} and Detfusion~\cite{sun2022detfusion} introduce the object detection model to drive the fusion model's ability to preserve semantic attributes, enabling foreground objects in the scene to be effectively highlighted in the fused image. MRFS~\cite{zhang2024mrfs} and RMR-Fusion~\cite{zhang2024robust} point out that these methods fail to genuinely inject semantic features into the fused image solely through optimization guidance. Thus, they deeply couple segmentation models with fusion models at the feature level and employ joint optimization, achieving further performance improvement. Notably, although the coupling of vision and semantics holds the potential for mutual enhancement, bridging the optimization gap and resolving conflicts between them remains an extremely complex and challenging task, as shown in Fig.~\ref{fig:1} (a). Therefore, this leads to compromised results in existing hybrid vision–semantic-driven methods.

\subsection{Consideration and Contribution}
Based on the above analysis, it is difficult to achieve a fused result that simultaneously preserves visual quality and semantic attributes through forced coupling. This leads us to a bold idea: \textit{is it possible to separately optimize visual and semantic aspects to achieve optimal solutions, and then combine their results \textbf{without compromising either}?} Naturally, the key challenge is transformed into determining how the semantic results should be presented and how they can be integrated with the visual results. We notice that adversarial attack techniques~\cite{li2019universal,yuan2024adaptive,qing2024detection} can effectively disrupt the scene’s semantic distribution by injecting imperceptible perturbations into the image, thereby causing failures in downstream detection and segmentation tasks. This offers a promising insight for achieving the aforementioned bold idea. Specifically, in the MMIF task, the semantic results can also in turn be viewed as an imperceptible map, and by overlaying them onto the fused image, the semantic attributes can be enhanced without affecting visual quality.

The above motivation leads us to propose a new \textbf{decoupled visual–semantic fusion paradigm}, deriving AD-SRF, a novel attack-defense semantic reinforcement framework for MMIF. It is worth noting that while some existing methods like A$^2$RNet~\cite{li2025a2rnet} leverage adversarial attack techniques to enhance the robustness of fused images against external malicious attacks, our AD-SRF fundamentally differs in its objective: we employ the attack-defense game as a proactive mechanism to mine and inject latent semantic attributes, thereby directly boosting the performance of downstream tasks. As depicted in Fig.~\ref{fig:1} (b), it significantly enhances the support for downstream semantic tasks, while fully preserving the visual quality of the fused image. Before all else, we establish a strong visual baseline for MMIF, involving a transformer-based restorative visual fusion (RVF) network. By leveraging the multi-modal complementarity, RVF introduces cross-modal retrieval (CMR) modules to purify and integrate features, generating clear fused images even under challenging conditions such as low light, noise, and color distortion, with the guidance of the information restoration-fusion collaborative loss. Building on it, we further introduce the attack-defense game (ADG) mechanism, which consists of an attacker and a defender. The attacker aims to produce a semantic perturbation map and overlay it onto the fused image to maximally disrupt semantic segmentation and render the defender's defenses ineffective. Oppositely, the defender is responsible for generating an imperceptible semantic map and superimposing it on the fused image to resist the perturbation map imposed by the attacker while maximizing the improvement in semantic segmentation performance. Through the continuous evolution and ongoing interplay of the attacker and defender, a high-quality imperceptible semantic map can be obtained, which can be considered to encapsulate the complete semantic distribution required for scene understanding. Finally, by combining the base fused image with an imperceptible semantic map, we can obtain a fused result that possesses both visual quality and semantic attributes, as presented in Fig.~\ref{fig:1} (c).

Our contributions can be summarized as follows:
\begin{itemize}
	\item We propose a new decoupled visual–semantic fusion paradigm, deriving an attack-defense semantic reinforcement framework for MMIF. To our knowledge, this is the first to achieve lossless decoupling of visual and semantic optimization using attack-defense technologies, overcoming the performance bottleneck of existing methods in balancing visual quality and semantic attributes.
	\item A restorative visual fusion network is designed to establish a robust visual baseline, using CMR modules and information restoration-fusion collaborative loss to generate clean fused images in degraded environments.
	\item We develop an attack-defense game mechanism to mine the scene's semantic distribution and customize it into an imperceptible map, enabling semantic attribute injection without compromising the fused image's visual quality.
	\item Extensive experiments demonstrate that our AD-SRF not only achieves high-quality visual representation under challenging conditions, but also significantly improves the accuracy of semantic segmentation.
\end{itemize}

\begin{figure*}[t]
	\centering
	\includegraphics[width=1\linewidth]{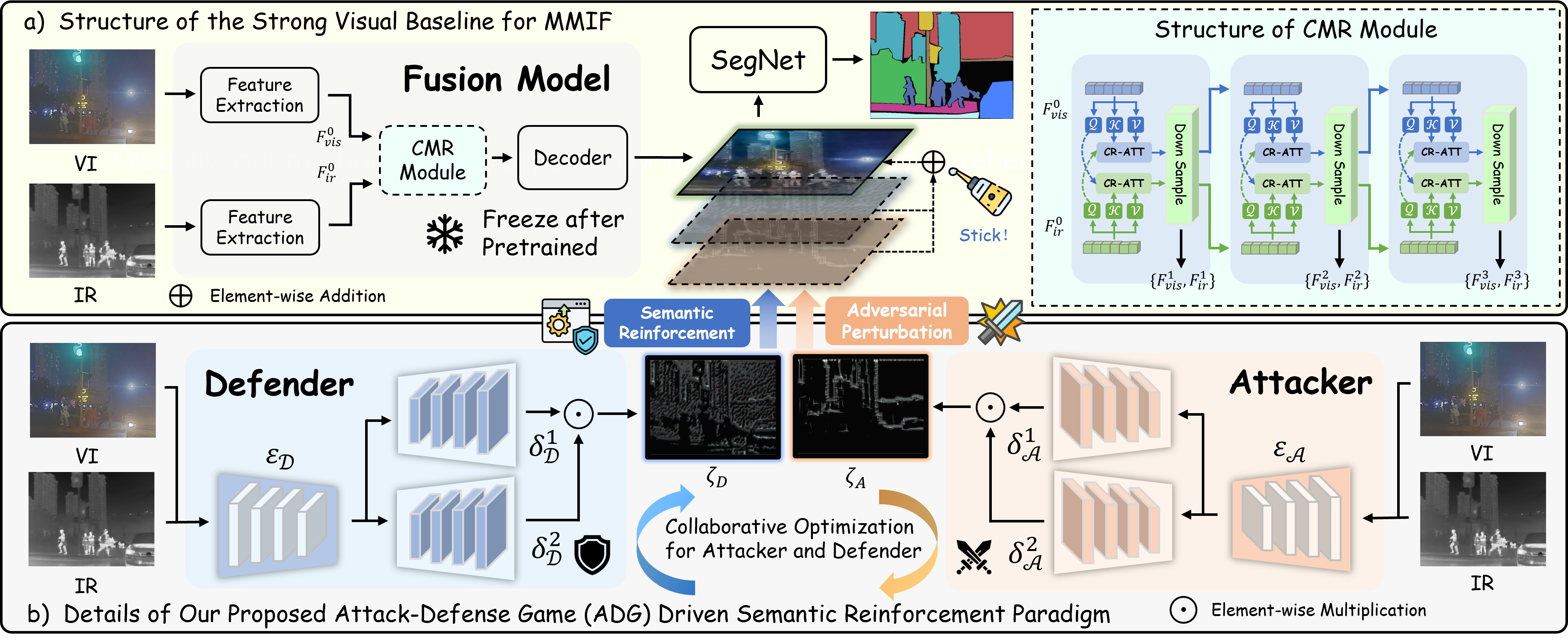}
	\vspace{-0.2in}
	\caption{The overall pipeline of our proposed attack-defense semantic reinforcement framework for multi-modal image fusion (AD-SRF). It starts from a strong visual baseline and leverages an attack-defense game to learn imperceptible semantic maps, improving the visual quality of fused images and their semantic affinity for the downstream segmentation task under degraded conditions.}\label{fig:2}
\end{figure*}

\section{Methodology}
\subsection{Problem Formulation}
In real-world scenarios with complex degradations, MMIF aims to reconstruct a single highly representative fused image $I_{f}$ by integrating complementary information from multi-modal images $I_{src}=\{I_{vis}, I_{ir}\}$, where $I_{vis}$ and $I_{ir}$ indicate the visible and infrared modalities, respectively. An ideal fused image is expected to satisfy two objectives simultaneously: preserving high-fidelity visual details for human observation while maintaining discriminative semantic attributes for accurate machine decision-making. Formally, the above goals can be achieved by minimizing a joint objective function:
\begin{equation}
	\min_{\theta_f} \left[ \mathcal{L}_{v}\big(\Phi(I_{src}; \theta_f)\big) + \lambda \mathcal{L}_{s}\big(\Psi(\Phi(I_{src}; \theta_f))\big) \right], \label{eq:1}
\end{equation}
where $\Phi(\cdot; \theta_f)$ denotes the fusion network parameterized by $\theta_f$, $\mathcal{L}_{v}$ measures the visual reconstruction fidelity, and $\mathcal{L}_{s}$ assess the semantic consistency via a downstream segmentation network $\Psi$. However, as previously analyzed, simultaneously optimizing these two terms is non-trivial, since their required feature representations and optimization directions are not fully aligned. As a result, enforcing a strict coupling between them often results in suboptimal compromises.

To address this challenge that has long perplexed the community, our AD-SRF proposes a novel decoupled visual–semantic fusion paradigm that losslessly preserves the visual quality of the fused results while ensuring its semantic affinity for downstream decision-making tasks. Its core idea is to employ an attack-defense game to decouple semantic attributes from visual representations, enabling each to be individually optimized to its best before recombination. The derived attack-defense semantic reinforcement framework for MMIF, AD-SRF, is demonstrated in Fig.~\ref{fig:2}. As seen, AD-SRF decouples Eq.~\eqref{eq:1} into two distinct, sequential optimization phases: \textit{\textbf{visual baseline}} and \textit{\textbf{semantic reinforcement}}.

The first phase focuses on building a strong visual fusion network to satisfy the demands of humans' observation, which also serves as the visual baseline for the second phase. In our view, a high-quality fused image should not only preserve salient information from the source sensors but also possess robust restoration capabilities to adapt to the adverse conditions of real-world scenarios. Motivated by this, we posit that multi-modal image fusion and restoration are not disjoint endeavors but are characterized by profound intrinsic complementarity and the potential for mutual reinforcement. On the one hand, the cross-modal feature extraction and interaction inherent in the fusion process enable a synergistic integration of complementary features, offering vital cross-modal cues to assist in image restoration. On the other hand, effective degradation removal enhances the clarity and fidelity of features, thereby facilitating more accurate information fusion. Leveraging this integrated perspective, we formalize the first-stage visual task as: ${I}{^H_f} = \Phi(I^{D}_{vis}, I^{D}_{ir};\theta_f)$, where $\Phi$ indicates the transformer-based restorative visual fusion (RVF) network, and the superscripts $H$ and $D$ denote high-quality and degraded, respectively. The realization of this restoration-fusion synergy necessitates the joint contribution of network architecture and objective constraints. To this end, we first introduce the cross-modal retrieval (CMR) modules to retrieve complementary
features required for restoration while simultaneously performing information fusion, continuously purifying and integrating multi-modal features. Furthermore, the information restoration–fusion collaborative loss incorporates clean multi-modal images as references into the fusion objective, guiding the RVF network toward robust restoration-fusion. Overall, the build of the first-stage visual baseline is formulated as follows: 
\begin{equation}
	\label{eq:2}
	\min_{\theta_f} \left[ \mathcal{L}_{f}( \Phi(I^{D}_{vis}, I^{D}_{ir};\theta_f) \rightarrow \{I^{H}_{vis}, I^{H}_{ir}\})\right].
\end{equation}
By solving this, the visually pleasing images ${I}{^H_f}$ are obtained even under harsh conditions, which are then frozen to serve as a robust visual baseline for the subsequent semantic optimization phase.

The second phase focuses on boosting semantic performance while maintaining the high visual fidelity of the first phase. Inspired by adversarial attack techniques, we posit that if meticulously crafted perturbations can significantly degrade decision accuracy by mining model vulnerabilities, this destructive logic can naturally be transformed into a constructive driving force. We believe that by utilizing adversarial attacks to pinpoint fragile semantic regions within fused images and using them to induce the training of a restorative defender, the network can be compelled to capture more robust, intrinsic features through the adversarial interplay. Thereby, we freeze the RVF network $\Phi$ obtained from the first phase and develop the attack-defense game (ADG) mechanism to mine and integrate semantic attributes into ${I}{^H_f}$ without compromising its visual quality. 
Specifically, this process involves a min-max game between an attacker and a defender. The attacker $\mathcal{A}$ aims to generate a perturbation $\zeta_A$ that maximizes the segmentation loss to identify semantic vulnerabilities: $\zeta_{A}$= $\mathcal{A}(I^{D}_{vis}, I^{D}_{ir};\theta_A)$. By superimposing $\zeta_{A}$ onto ${I}{^H_f}$, the image’s semantic attributes are corrupted, causing downstream segmentation to fail. In contrast, we introduce a defender $\mathcal{D}$ to generate an imperceptible semantic map $\zeta_D$ to counteract the attack and reinforce semantic features: $\zeta_{D}$= $\mathcal{D}(I^{D}_{vis}, I^{D}_{ir};\theta_D)$. By overlaying $\zeta_{D}$ onto ${I}{^H_f}$, the semantic attributes of the fused image are enhanced, thereby improving the performance of downstream segmentation. To sum up, the joint optimization problem of our ADG mechanism is formulated as:
\begin{equation}
	\label{eq:minmax_game} 
	\min_{\theta_D} \max_{\theta_A} \left[ \mathcal{L}_{s}\Big(\Psi\big( {I}{^H_f} + \zeta_D + \zeta_A \big)\Big)\right].
\end{equation}
It is worth noting that to strictly preserve the visual appearance, we enforce a constraint on the magnitude of $\zeta$, ensuring that the injected semantic cues remain visually imperceptible. 

The essence of this decoupling lies in the spatial and parametric separation of conflicting objectives. By freezing the RVF network $\Phi$, we establish an immutable visual baseline that acts as a safeguard against catastrophic degradation of visual features during adversarial optimization. Instead of re-modifying the pixel distributions, our paradigm treats semantic reinforcement as an additive, imperceptible semantic bias $\zeta_D$. By enforcing a strict constraint on its magnitude, we ensure that the visual integrity of the fused image remains intact. This mechanism ensures that critical semantic cues are injected into the feature space without perturbing the macroscopic visual appearance, effectively resolving the inherent trade-off between visual clarity and semantic precision found in traditional coupled architectures. Ultimately, this decoupling strategy enables AD-SRF to achieve a synergetic \textbf{win-win} outcome, where visual fidelity is rigorously anchored by the fixed baseline while semantic intelligence is adaptively augmented via the ADG mechanism.

\subsection{The Restorative Visual Fusion (RVF) Baseline}\label{sec:network}
In multimodal vision tasks, image restoration and fusion are inseparable processes that are deeply interconnected and mutually dependent. This interdependence is manifested in a synergistic relationship: the restoration process purifies feature representations and suppresses noise interference to benefit fusion, whereas cross-modal interactions provide critical information gains to facilitate robust restoration. Together, they effectively surmount the performance bottlenecks inherent in isolated tasks. To address these intricately coupled challenges within a unified framework, we implement meticulous designs across both \textbf{architectural} and \textbf{constraint} dimensions.

\textbf{Architectural Configuration.} To effectively address the multiple demands of image restoration and multimodal fusion, such as feature denoising, information completion and high-quality reconstruction, the RVF network is designed as a systematic architecture consisting of three key stages: feature extraction, feature retrieval, and feature decoding. This progressive design enables a smooth transition from raw degraded signals to purified, cross-modal enhanced features, and ultimately to high-fidelity image reconstruction. First, since image restoration requires a holistic understanding of degraded scenes, while fusion tasks rely on capturing globally salient targets, we argue that extracting global contextual information is crucial for achieving robust and high-quality restoration–fusion results. Therefore, we employ a hybrid convolutional stem consisting of a convolutional layer followed by stacked MV2 blocks~\cite{mehta2022mobilevit} to extract robust initial representations from the degraded source images $I^{D}_{vis}, I^{D}_{ir} \in \mathbb{R}^{H \times W \times C}$, formulated as:
\begin{equation}
	F_{vis}^0 = \operatorname{MV2}(\operatorname{Conv}(I^{D}_{vis})), \quad F_{ir}^0 = \operatorname{MV2}(\operatorname{Conv}(I^{D}_{ir})).
\end{equation}
With these robust embeddings, we subsequently address the coupling between restoration and fusion by devising the CMR module. At the heart of the CMR module lies the premise that high-quality restoration-fusion performance hinges on the synergy between cross-modal and multi-scale feature interactions. To facilitate this retrieval and interaction process, the input features are first tokenized and projected into their corresponding Query ($Q$), Key ($K$), and Value ($V$) subspaces via linear transformations $\mathcal{F}^{qkv}$:
\begin{equation}
	\begin{split}
		\{Q_{vis}^{l}, K_{vis}^{l}, V_{vis}^{l}\} &= \mathcal{F}^{qkv}_{vis}(F_{vis}^{l-1}), \\
		\{Q_{ir}^{l}, K_{ir}^{l}, V_{ir}^{l}\} &= \mathcal{F}^{qkv}_{ir}(F_{ir}^{l-1}).
	\end{split}
\end{equation}

After that, we execute a bidirectional retrieval strategy to mutually reinforce feature representations by integrating complementary cues from the cross-modal context. Specifically, $Q_{vis}^{l}$ queries the infrared context $\{K_{ir}^{l}, V_{ir}^{l}\}$ to replenish geometric details lost in degraded visible regions, leveraging the structural information of infrared images as a cross-modal prior for visible restoration. Simultaneously, $Q_{ir}^{l}$ retrieves visible context to incorporate rich textures, compensating for the sparse environmental representation of infrared images. Rather than simple feature superposition, this retrieval process leverages effective features from the peer modality to achieve real-time purification and calibration of degraded representations within the current modality. In this manner, an organic unification is achieved between the cross-modal interaction required for fusion and the visual cues vital for restoration. What's more, to ensure the feature purity and effectiveness of each modality, the retrieved information is aggregated via residual connections to preserve intrinsic modality characteristics while reinforcing them with cross-modal complements:
\begin{equation}
	\label{eq:retrieval}
	\begin{split}
		F_{vis}^{l} &= \text{Softmax}\left(\frac{Q_{vis}^{l}K_{ir}^{l}}{\sqrt{d_k}}\right) V_{ir} + F_{vis}^{l-1}, \\
		{F}_{ir}^{l} &= \text{Softmax}\left(\frac{Q_{ir}^{l}K_{vis}^{l}}{\sqrt{d_k}}\right) V_{vis} + F_{ir}^{l-1},
	\end{split}
\end{equation}
where $d_{k}$ is the scaling factor. Subsequent to the resolution of feature interaction, in order to enlarge the receptive field and satisfy the demand for multi-scale representations, the refined features undergo a downsampling operation $\mathcal{S}(\cdot)$ after each block: $F_{vis}^{l}, F_{ir}^{l} \gets \mathcal{S}(F_{vis}^{l}), \mathcal{S}(F_{ir}^{l})$, where $l\in\{1, 2, 3\}$. Finally, the feature pyramid $\{F_{vis}^{l}, F_{ir}^{l}\}_{l=1}^{3}$ is aligned and concatenated then fed into the decoder to reconstruct $I_f^H$, effectively unifying high-fidelity restoration-fusion within a single high-quality output.

\textbf{Constraint Formulation.}
An ideal fused representation should transcend simple pixel-level merging; it must simultaneously mitigate environmental degradations and integrate the most salient features from source modalities to achieve superior visual fidelity. Specifically, an excellent fused image is expected to possess sharp structural textures, faithful color consistency, and distinct global contrast, ensuring it is both visually pleasing for human observation and highly informative for downstream machine perception. To instantiate these objectives, we maintain that the generated high-quality image $I_f^H$ should be rigorously constrained to align with the reference target across these three primary dimensions. Therefore, the proposed  information restoration-fusion collaborative loss $\mathcal{L}_{f}$ can be formulated as: $\mathcal{L}_{f} = \mathcal{L}_{str} + \mathcal{L}_{col} + \mathcal{L}_{con}$.

Regarding structural fidelity, clearly distinguishable structural features are highly beneficial for both human perception and downstream machine tasks. To drive the RVF network to reconstruct sharp geometric boundaries from degraded inputs, $\mathcal{L}_{str}$ aligns the gradient of $I_f^H$ with the target sharpened gradients in both horizontal ($\nabla_x$) and vertical ($\nabla_y$) directions:
\begin{equation}
	\label{eq:str_loss}
	\begin{split}
		\mathcal{L}_{str} &= \| \nabla I_{f}^H - \max(\nabla I^{H}_{vis}, \nabla I^{H}_{ir}) \|_1, \\
	\end{split}
\end{equation}
where $\nabla I = |\nabla_x I| + |\nabla_y I|$.
Regarding chromatic quality, $\mathcal{L}_{col}$ is introduced to drive the reconstruction of a perceptually natural appearance by aligning the chrominance distribution in the YCbCr space with the high-quality target:
\begin{equation}
	\mathcal{L}_{col} = \| Cr(I_{f}^H) - Cr(I^{H}_{vis}) \|_1 + \| Cb(I_{f}^H) - Cb(I^{H}_{vis}) \|_1.
\end{equation}
This constraint ensures that the network effectively recovers realistic hues from degraded inputs, achieving a seamless integration of image restoration and multi-modal fusion to produce a visually authentic representation.

Besides, the preservation of intensity or contrast consistency remains indispensable for the reconstruction of $I_f^H$. We argue that an ideal fused result must strike a balance between background textural richness and foreground target saliency. Nevertheless, existing approaches typically impose global, homogeneous constraints, failing to account for the regional diversity of required fusion strategies. This oversight serves as a primary bottleneck hindering superior visual quality. Therefore, 
we rethink the constraint from a \textit{pixel-wise} alignment to a \textit{region-wise} adaptive preservation perspective and propose the contrastive loss $\mathcal{L}_{con}$ to guide the network toward content-adaptive fusion: $\mathcal{L}_{con} = \mathcal{L}_{tar} + \mathcal{L}_{bkg}$. In this manner, $\mathcal{L}_{con}$ enables the distinct processing of foreground and background through specialized strategies. Specifically, we utilize a binary mask $\mathcal{M}$ to spatially partition the image, where $\mathcal{M}=1$ denotes the foreground target regions and $\mathcal{M}=0$ represents the background.
For target regions where saliency is paramount, we adopt a maximization strategy to drive the network toward learning and reconstructing the most distinctive features from the high-quality target:
\begin{equation}
	\mathcal{L}_{tar} = \| \mathcal{M} \odot I_{f}^H - \mathcal{M} \odot \max(I^{H}_{vis}, I^{H}_{ir}) \|_1,
\end{equation}
where $\odot$ denotes Hadamard product and $\|\cdot\|_1$ is the $l_1$-norm.
Conversely, for background regions, the primary goal is to preserve rich environmental textures. Thus, we employ an adaptive weighting strategy based on local information richness. By calculating the standard deviations $\{\sigma_{vis}, \sigma_{ir}\}$ as indicators of texture density, we dynamically assign weights to drive the network to focus on the modality with richer details. The adaptive weights for each modality are computed as:
\begin{equation} 
	w_{m} = \frac{\sigma_{m}}{\sum_{k}^ {\{vis, ir\}} \sigma_{k} + \epsilon}, 
\end{equation}
where $m \!\in \!\{vis, ir\}$, $\epsilon$ is a small constant for numerical stability. Guided by these texture-aware priors, we formulate the background loss to minimize the weighted reconstruction discrepancy, ensuring background textural richness:
\begin{equation}
    \mathcal{L}_{bkg} = \!\! \sum_{m}^{\{vis, ir\}} \!\!  w_{m} \cdot \|(1-\mathcal{M}) \odot I_{f}^H - (1-\mathcal{M}) \odot I^{H}_{m}\|_{1}.
\end{equation}
By combining a target prioritization mechanism for salient regions with information richness weights for textural backgrounds, $\mathcal{L}_{con}$ empowers the network to function as an intelligent content arbiter, autonomously integrating the valuable cues from distinct modalities into a coherent representation.

Collectively, the proposed restoration-fusion visual baseline establishes a strong foundation for both human vision and high-level tasks. Through the joint efforts of the RVF network's architectural design and loss constraints, $I_{f}^H$ exhibits a unique duality of excellence, combining the degradation-free clarity of restoration with the comprehensive information richness of fusion. This superior visual quality serves as a robust anchor, ensuring the downstream decision-making is based on images that is both perceptually natural and semantically trustworthy.

\subsection{Adversarial Semantic Reinforcement}
Building upon the high-fidelity visual baseline established in the first stage, we proceed to the second phase: \textit{Adversarial Semantic Reinforcement}. In order to strictly decouple semantic optimization from visual restoration, we freeze the parameters of the RVF baseline, treating the generated high-quality fused image $I_f^H$ as a stable anchor. However, the restoration-fusion mapping process $\Phi: \{I_{vis}^D, I_{ir}^D\} \to I_f^H$ inevitably functions as a lossy compression operator. During this projection, fine-grained semantic features (e.g., texture structures or edge details) are prone to be attenuated or discarded along with the noise. Consequently, breaking the semantic bottleneck by replenishing the latent feature information lost during the $\Phi$ transformation process has emerged as a critical necessity. 

We draw inspiration from the structural sparsity paradigm~\cite{CVPR24_EGS_TSSA} utilized in adversarial attacks. While these methods are primarily known for their ability to synthesize subtle perturbations that effectively sabotage downstream performance, we posit that this generative capacity can be repurposed. By identifying and targeting the same semantic blind spots that an attacker would exploit, the mechanism can be redirected to semantic reinforcement. With this aim, we design an \textit{Attack-Defense Game (ADG)} mechanism to explicitly embed robust semantic attributes into the frozen visual baseline. This mechanism relies on the competition between two generators: an attacker $\mathcal{A}$ (parameterized by $\theta_A$) and a defender $\mathcal{D}$ (parameterized by $\theta_D$). Specifically, both agents employ an architecture composed of a ResNet-based~\cite{he2016identity} shared encoder $\varepsilon $ and dual-branch decoders $\delta_1$ and $\delta_2$. This architecture structurally decouples the prediction of spatial locations from the estimation of perturbation magnitudes, enabling independent constraints over each attribute. Particularly, the deep multi-modal features $\mathcal{F} = \varepsilon(I_{vis}^D, I_{ir}^D)$ are routed into two parallel branches. The first branch predicts the position mask $\mathbf{M}$ while the second branch estimates the raw intensity, which is constrained via a scaled activation. The generation process is formulated as:
\begin{equation}
	\mathbf{M} = \delta_1(\mathcal{F}), \quad \mathbf{A} = \tau \cdot \tanh\Big(\delta_2(\mathcal{F})\Big),
\end{equation}
here, $\mathbf{M}$ denotes the position mask (indicating \textit{where} to modify), while $\mathbf{A}$ represents the pixel-wise intensity map (determining \textit{how much} to modify). In this formulation, $\tanh(\cdot)$ maps the raw decoder outputs to the interval $(-1, 1)$, and $\tau$ serves as a hard scaling factor to strictly confine the final intensity values within $[-\tau, \tau]$. By leveraging this decoupling strategy, we can enforce sparsity constraints on $\mathbf{M}$ via an $l_1$-norm sparsity loss, while strictly confining the magnitude of $\mathbf{A}$ within a hyperparameter $\tau$ to control its range. The final semantic perturbation/reinforcement map generated by $\mathcal{A}$/$\mathcal{D}$ is synthesized via the Hadamard product: $\zeta = \mathbf{M} \odot \mathbf{A}$. These ensure that the generated maps are both sparsely distributed and numerically controlled, thereby achieving the optimal balance between semantic effectiveness and visual imperceptibility.
		
Facilitated by this precise generation capability, ADG is realized through the antagonistic interaction between the two agents. In the following, we will first elaborate on the distinct optimization objectives of $\mathcal{A}$  and $\mathcal{D}$ , and then present the unified Min-Max training paradigm that drives their competition.
		
\textbf{Attacker: Semantic Vulnerability Mining.}
In the realm of digital image processing, adversarial attacks are formally modeled as an optimization problem aimed at discovering the worst-case perturbation within a constrained neighborhood. Given a clean input $x$ and a target model $\Psi$, the objective is to find a perturbation $\zeta$ that maximizes the divergence from the ground truth, subject to a perceptual constraint $\|\zeta\|_p \le \epsilon$:
\begin{equation}
	\label{eq:adv_attack}
\zeta^* = \operatorname*{arg\,max}_{\|\zeta\|_p \le \epsilon} \mathcal{L}_{s}(\Psi(x + \zeta)).
\end{equation}
The fundamental goal of this process is to synthesize perturbations that actively subvert the semantic decision mechanism of $\Psi$. The resulting adversarial examples represent the data points closest to the decision boundary, revealing the inherent robustness blind spots in the model's feature representation. 

Bridging this theoretical formulation to our context, we postulate that if the fused image $I_f^H$ is easily subverted by such perturbations, it indicates that the underlying semantic features are fragile or insufficient. Therefore, we repurpose this offensive paradigm from a security threat into a diagnostic mechanism for \textit{Semantic Vulnerability Mining}. Unlike conventional gradient-based methods that perform an iterative gradient ascent to approximate the solution of Eq.~\eqref{eq:adv_attack} for each specific input, we parameterize this search process using the generator network $\mathcal{A}$ to ensure efficiency and transferability. $\mathcal{A}$ aims to learn a generalizable distribution of malicious perturbations $\zeta_A$ that can effectively trigger segmentation failures in the frozen baseline $I_f^H$ and actively penetrate the defense mechanisms of $\mathcal{D}$, while maintaining visual imperceptibility. Therefore, we reformulate Eq.~\eqref{eq:minmax_game} for $\mathcal{A}$ as:
\begin{equation}
\label{eq:A_loss}
\hspace{-0.68em}
\resizebox{0.92\linewidth}{!}{ 
$\displaystyle 
\min_{\theta_A} \!\left(\! - \Big[ \mathcal{L}_{s}(\Psi(I_f^H {+} \zeta_A)) {+} \mathcal{L}_{s}(\Psi(I_f^H {+} \zeta_A {+} \bar{\zeta_D})) \!\Big] {+} \mathcal{R}(\mathbf{M}_A) \!\right),$}
\end{equation}
where $\mathcal{R}(\mathbf{M}_A) = \max(0, \|\mathbf{M}_A\|_1 - \xi_{max}) + \max(0, \xi_{min} - \|\mathbf{M}_A\|_1)$ denotes the regularization term to constrain the position map $\mathbf{M}_A$ of $\zeta_A$. This optimization process orchestrates three distinct behaviors within the adversarial game. The first semantic term targets the intrinsic vulnerabilities of the baseline $I_f^H$ directly, ensuring $\mathcal{A}$ learns to identify fundamental semantic weaknesses independent of any interference. The second term enforces adaptive penetration by incorporating $\bar{\zeta_D}$, which is the frozen defender output from the previous alternating iteration. This acts as a counter-defense objective, compelling $\mathcal{A}$ to evolve dynamically and discover robust attacks that can break through the current defense mechanisms. Finally, the regularization term restricts the spatial footprint of $\mathbf{M}_A$ to the interval $[\xi_{min}, \xi_{max}]$. This constraint guarantees that the generated perturbations remain visually imperceptible while preventing the attack from vanishing.
		
\begin{figure}[t]  
\centering
\includegraphics[width=\columnwidth]{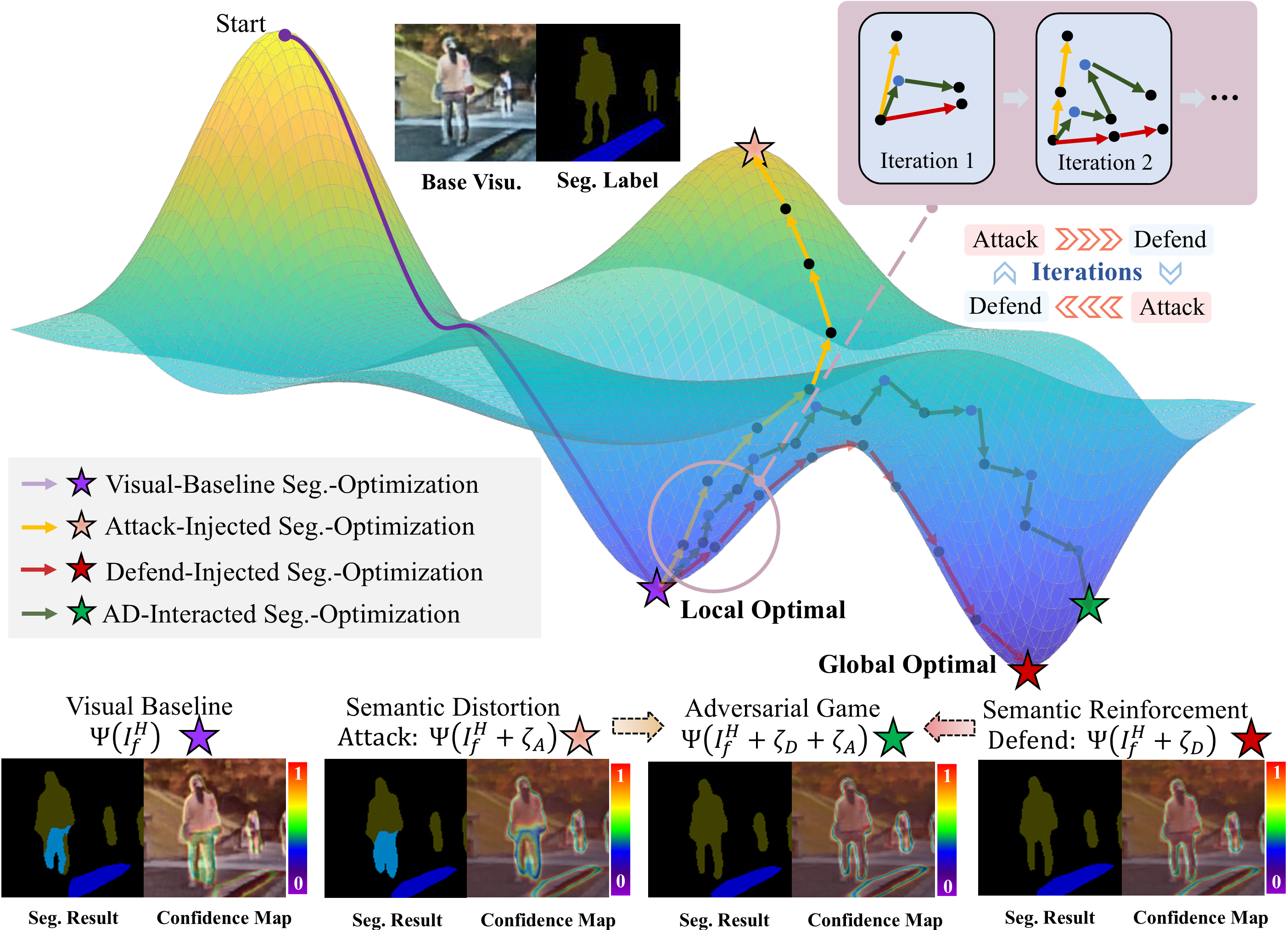} 
\caption{An Optimization Perspective on ADG: Principles and Visualizations. We demonstrate the principles of ADG from an optimization perspective, complemented by confidence map visualizations that verify the stability and precision of the reinforced features.} 
\label{fig:3}
\end{figure}
		
\textbf{Defender: Robust Semantic Reinforcement.}
By engaging $\mathcal{D}$ in an adversarial game against $\mathcal{A}$, we simulate a worst-case semantic environment where features are severely distorted. We posit that semantic reinforcement and adversarial defense are intrinsically unified by a shared goal: achieving semantic robustness through the active recovery of feature vulnerabilities. Specifically, reinforcement provides the foundation by identifying and restoring essential semantic cues that were lost or degraded. Adversarial defense takes this a further step requiring the network to not only replenish missing information but also actively neutralize malicious interference to ensure stable semantic performance. Thus, when $\mathcal{D}$ is rigorously trained to neutralize adversarial attacks, it can achieve superior precision when refining the clean visual baseline. 

Realizing this property, we explicitly position $\mathcal{D}$ to serve a dual function: acting simultaneously as a robust defender against potential attacks and a semantic enhancer for intrinsic image quality. To implement this, we employ the same decoupled generator architecture for $\mathcal{D}$, and reformulate its optimization objective as:
\begin{equation}
    \label{eq:D_loss}
    \hspace{-0.68em}
    \resizebox{0.92\linewidth}{!}{
        $\displaystyle 
        \min_{\theta_D} \!\left(\! \Big[ \mathcal{L}_{s}(\Psi(I_f^H {+} \zeta_D)) {+} \mathcal{L}_{s}(\Psi(I_f^H {+} \zeta_D {+} \bar{\zeta_A})) \!\Big] {+} \mathcal{R}(\mathbf{M}_D) \!\right),$}
\end{equation}
where $\bar{\zeta_A}$ denotes the fixed attacker output from the current iteration and $\mathcal{R}(\mathbf{M}_D)$ is the regularization term for the defender's position map $\mathbf{M}_D$ defined similarly as above. Mirroring $\mathcal{A}$'s structure, this optimization goal is also composed of three components. While the network is trained to handle the more challenging task of neutralizing adversarial attacks, the first term serves to prevent the defender $\mathcal{D}$ from overfitting to the specific defense of $\zeta_A$. By doing so, it ensures that the model maintains its fundamental capability to replenish attenuated semantic cues for $I_f^H$. The second term, conditioning on the attacked input $I_f^H {+} \bar{\zeta_A}$, serves as the adversarial robustness objective. It compels $\mathcal{D}$ to neutralize the semantic distortions introduced by $\mathcal{A}$, ensuring the defense is resilient to worst-case shifts. Finally, the regularization term strictly bounds $\mathbf{M}_D$ ensuring the defense mechanism operates as a precise correction rather than a global modification to preserve the visual fidelity of $I_f^H$.
		
\begin{algorithm}[t]
	\caption{Adversarial Semantic Reinforcement Training.}
	\label{alg:training_process}
	\begin{algorithmic}[1]
		\STATE \textbf{Input:} multi-modal dataset $\mathcal{S} = \{(I_{vis}^{D}, I_{ir}^{D}, y)\}$, frozen fusion baseline $\Phi$, pretrained segmentation network $\Psi$
		\STATE \textbf{Models:} Attacker ($\theta_A$), Defender ($\theta_D$), Segmentation Network ($\theta_\Psi$) (backbone frozen)
		\STATE \textbf{Initialize:} Attacker ($\theta_A$), Defender ($\theta_D$)
		\FOR{$t = 1$ \TO $T$}
		\STATE Sample a mini-batch in $\mathcal{S}$
		\STATE Generate frozen baseline $I_f^H = \Phi(I_{vis}^{(i)}, I_{ir}^{(i)})$.
		\STATE \textit{\% Phase I: Optimize Attacker}
		\STATE Freeze $\theta_D$ and $\theta_{\Psi}$
		\STATE Get $\bar{\zeta_D}$ (no grad) and $\zeta_A$ via $\mathcal{D}(\cdot)$ and $\mathcal{A}(\cdot)$
		\STATE Calculate $\mathcal{L}_{A}$ via Eq.~\eqref{eq:A_loss}
		\STATE Update $\theta_A \leftarrow \theta_A - \eta \nabla_{\theta_A} \mathcal{L}_{A}$
		
		\STATE \textit{\% Phase II: Optimize Defender and Finetune SegNet}
		\STATE Freeze $\theta_A$
		\STATE Get $\bar{\zeta_A}$ (no grad) and $\zeta_D$ via $\mathcal{A}(\cdot)$ and $\mathcal{D}(\cdot)$
		\STATE Calculate $\mathcal{L}_{D}$ via Eq.~\eqref{eq:D_loss}
		\STATE Update Defender: $\theta_D \leftarrow \theta_D - \eta \nabla_{\theta_D} \mathcal{L}_{D}$
		\STATE Finetuning SegNet: $\theta_{\Psi} \leftarrow \theta_{\Psi} - \gamma \cdot \eta \nabla_{\theta_{\Psi}} \mathcal{L}_{D}$
		\ENDFOR
		\STATE \textbf{Return:} Optimized Defender $\theta_D^*$ and Adapted SegNet $\Psi^*$.
	\end{algorithmic}
\end{algorithm}
		
\textbf{Optimization Strategy: Adversarial Training.}
To effectively solve the proposed Attack-Defense Game (ADG), we adopt an adversarial training strategy based on alternating optimization within each mini-batch, as illustrated in Fig.~\ref{fig:3} and Algorithm~\ref{alg:training_process}. Specifically, to ensure that $\mathcal{A}$ effectively mines semantic vulnerabilities while $\mathcal{D}$ and $\Psi$ are synchronously optimized to counteract these threats, each training iteration is structured into two sequential phases: an independent attack phase followed by a joint defense and task adaptation phase.

\textit{Phase I: Attack Optimization.} We freeze  $\mathcal{D}$ and $\Psi$ to optimize $\mathcal{A}$. The objective of $\mathcal{A}$ here is twofold: to maximize the segmentation error on the current baseline and to penetrate the existing defense mechanism. By treating the previous defender output $\bar{\zeta_D}$ as a static context, $\mathcal{A}$ is forced to find the blind spots in the current defense configuration, generating perturbations $\zeta_A$ that can subvert the protected image. \textit{Phase II: Joint Defense and Task Adaptation.} We freeze $\mathcal{A}$ and switch to optimize the defense side. Here, $\mathcal{D}$ aims to neutralize the fixed attack $\bar{\zeta_A}$ and simultaneously enhance the semantic quality of the baseline. Crucially, we jointly optimize the segmentation network $\Psi$ in this phase to align it with the reinforced image distribution at the same time. However, performing full-parameter fine-tuning on these dynamic inputs is unfeasible, as it not only risks destabilizing the pre-trained feature space but also imposes prohibitive computational and memory costs. Therefore, we freeze the feature extraction backbone of $\Psi$ and update only its auxiliary head and decoder. To further ensure stability, we apply a scaling multiplier $\gamma$ to the learning rate of the segmentation network. This conservative update strategy allows the task head to adapt to the minute feature shifts caused by $\mathcal{D}$'s enhancement, without disrupting the robust semantic representations learned during pre-training. This two-phase cycle is executed cyclically, returning to Phase I after each joint update until the adversarial interaction achieves a stable and semantically robust convergence. After the adversarial training concludes, the optimal segmentation performance is achieved by $\Psi(I_f^H + \zeta_D)$. 

In summary, this antagonistic interplay enables the ADG mechanism to identify and reinforce semantic blind spots within the fused representation. By forcing the defender $\mathcal{D}$ to neutralize targeted attacks from $\mathcal{A}$, the network learns to replenish critical feature cues that are typically degraded by static fusion paradigms. Furthermore, the intensity-position decoupling ensures that these semantic updates remain visually imperceptible, allowing the model to achieve a significant leap in semantic performance while strictly preserving the high-fidelity appearance established in previous stages. This adversarial optimization ultimately ensures that the fused features are not only visually clean but also semantically dense and resilient for complex downstream tasks.

	\section{Experiments and Results}

\subsection{Implementation Details}

\textbf{Datasets.} To evaluate the robustness of AD-SRF across diverse degradation scenarios and substantiate its superiority in semantic enhancement, we selected several benchmarking datasets covering various challenges. MFNet \cite{ha2017mfnet} and FMB \cite{liu2023multi} Datasets are primarily utilized to assess visual performance under real-world degradations (e.g., low-light, fog, and rain) and to validate the semantic gains driven by the ADG mechanism. Specifically, MFNet comprises 784 training pairs and 393 testing pairs, while FMB consists of 1,220 training and 280 testing pairs. Additionally, the Polarization Image Dataset \cite{qiu2019polarization} and the RGB-NIR Scene Dataset \cite{brown2011multi} were utilized for zero-shot generalization experiments, providing a rigorous validation of AD-SRF's cross-modal universality and its capacity to generalize to novel sensing modalities.

\textbf{Training Details.}
The training protocol for AD-SRF is meticulously divided into three sequential phases, starting with the optimization of the RVF network to construct a robust visual baseline that serves both human aesthetic standards and subsequent perception tasks. Subsequently, we pre-train the segmentation network which adopts OverLock-B\cite{lou2025overlock} as backbone with fused images yielded by the frozen RVF baseline. Building on these foundations, the third stage introduces the Attack-Defense Game (ADG) training, where the framework undergoes joint adversarial optimization to explicitly mine and reinforce latent semantic attributes through an antagonistic interplay. All network modules are optimized via the AdamW optimizer with an initial learning rate of $6 \times 10^{-5}$, while the fine-tuning of the segmentation network in the final stage utilizes a learning rate multiplier of $\gamma = 10^{-2}$ to maintain stable and effective task adaptation. All experiments are conducted on four NVIDIA GeForce RTX 3090 GPUs with 24GB memory and Intel(R) Xeon(R) Platinum 8375C CPU.

\textbf{Evaluation Strategy.} We employ eleven state-of-the-art image fusion methods for comparison, including SeAFusion \cite{tang2022image}, MRFS \cite{zhang2024mrfs}, EMMA \cite{zhao2024equivariant}, KDFuse \cite{yang2025kdfuse}, Text-IF \cite{yi2024text}, A$^2$RNet~\cite{li2025a2rnet}, TDFusion \cite{bai2025task}, SAGE \cite{wu2025every}, URFusion \cite{xu2025urfusion}, OmniFuse \cite{zhang2025omnifuse} and ControlFusion \cite{tang2025controlfusion}. For visual quality assessment, six objective metrics are used, including PSNR, AG, EN, SSIM, $Q^{AB/F}$ and $N^{AB/F}$. For semantic segmentation evaluation, beyond the eleven aforementioned fusion methods, we also compare three state-of-the-art RGB-T segmentation algorithms including StitchFusion \cite{li2025stitchfusion}, MiLNet \cite{liu2025milnet} and KTB \cite{cai2025keep}. For these fusion methods' semantic comparison, we retrained the segmentation network on the generated fused images. The mean Intersection over Union (mIoU) metric is employed to quantify segmentation accuracy.

\begin{figure*}[htbp]
	\centering
	\includegraphics[width=1\linewidth]{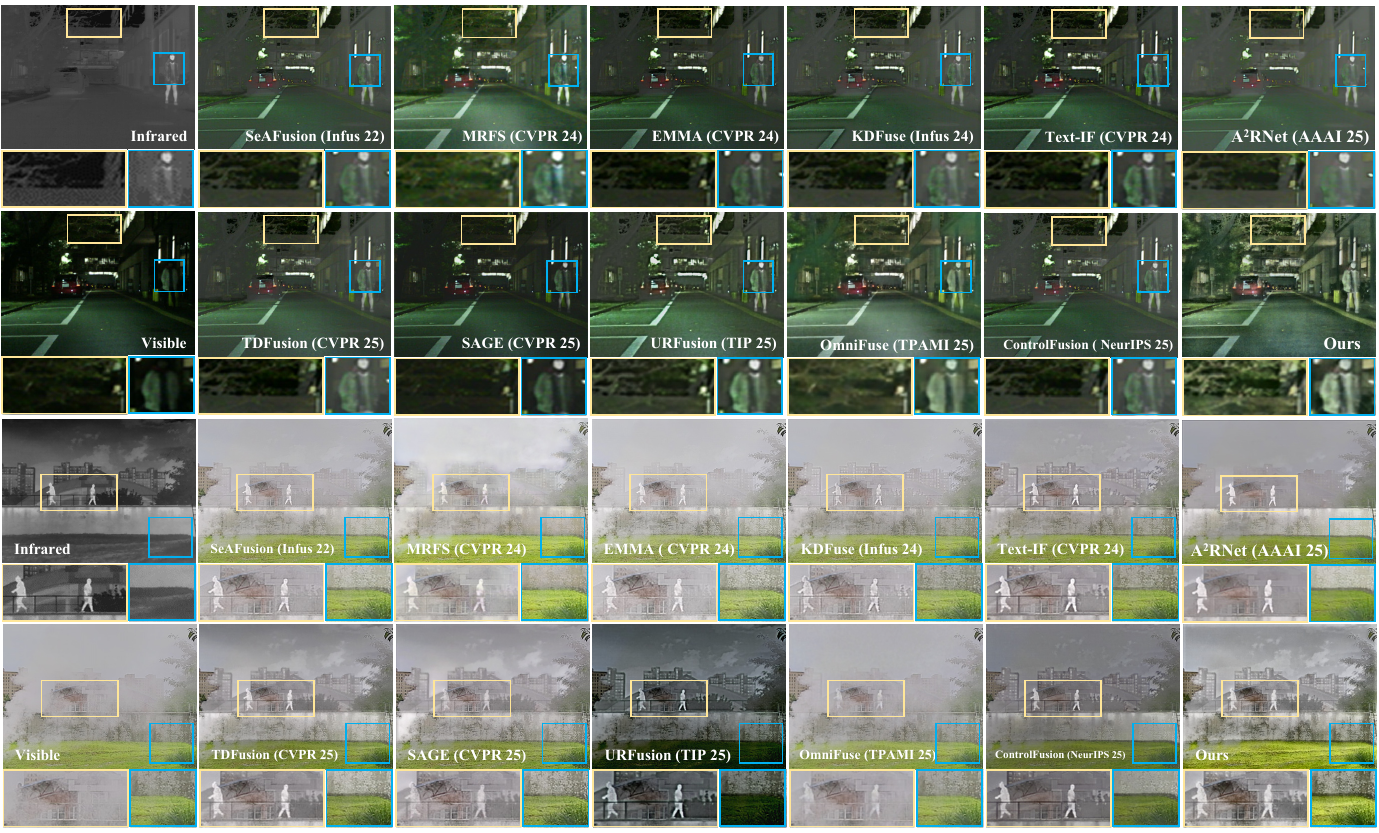}
	\vspace{-0.2in}
	\caption{Qualitative comparison of our method and existing image fusion methods under standard setting.}\label{fig:4}
\end{figure*}

\begin{table*}[t]
\vspace{-1pt}
\centering
\caption{Quantitative vision results on MFNet and FMB datasets under standard setting.}
\label{tab:MetricComp}
\vspace{-2pt}
\renewcommand{\arraystretch}{1.2} 
\setlength{\tabcolsep}{4pt} 
\begin{tabular}{l|cccccc|cccccc} 
\toprule
\multirow{2}{*}[-3pt]{\textbf{Methods}} & \multicolumn{6}{c|}{\textbf{MFNet}} & \multicolumn{6}{c}{\textbf{FMB}} \\
\cline{2-13}
& \rule{0pt}{11pt} \textbf{PSNR} $\uparrow$ & \textbf{AG} $\uparrow$  & \textbf{EN} $\uparrow$  & \textbf{SSIM} $\uparrow$  & $\bm{Q^{AB/F}}$ $\uparrow$  & $\bm{N^{AB/F}}$ $\downarrow$  & \textbf{PSNR} $\uparrow$  & \textbf{AG} $\uparrow$  & \textbf{EN} $\uparrow$  & \textbf{SSIM} $\uparrow$  & $\bm{Q^{AB/F}}$ $\uparrow$  & $\bm{N^{AB/F}}$ $\downarrow$  \\ 
\midrule 
SeAFusion\cite{tang2022image}   & 63.8112 & 3.6301 & 6.4217 & 0.3979 & 0.4700 & 0.0225 & 59.9176 & 4.2748 & 6.7538 & 0.2829 & 0.2685 & 0.0107 \\
MRFS \cite{zhang2024mrfs}       & 62.7057 & 3.2083 & \cellcolor{myLightBlue}7.0467 & 0.3595 & 0.4061 & 0.0207 & 60.5819 & 3.6545 & 6.7777 & 0.2593 & 0.2219 & 0.0122 \\
EMMA \cite{zhao2024equivariant} & 63.3372 & 4.0151 & 6.6118 & 0.3881 & 0.4896 & 0.0219 & 60.8850 & 4.6881 & 6.7729 & \cellcolor{myLightBlue}0.2968 & 0.3333 & 0.0110 \\
KDFuse \cite{yang2025kdfuse}    & 63.7641 & 3.7307 & 6.4990 & 0.3840 & 0.5067 & 0.0224 & 60.2450 & 4.2054 & 6.6484 & 0.2676 & 0.2662 & 0.0114 \\
Text-IF \cite{yi2024text}       & 63.2369 & 4.0099 & 6.7773 & 0.4119 & \cellcolor{myLightRed}\textbf{0.5598} & 0.0205 & 60.8934 & 4.4739 & 6.7402 & 0.2784 & 0.3085 & 0.0097 \\
A$^2$RNet \cite{li2025a2rnet}       & 64.2232 & 2.4955 & 6.2564 & 0.2823 & 0.2565 & 0.0241 & 60.5301 & 2.6420 & 6.5280 & 0.1841 & 0.1257 & 0.0113 \\
TDFusion \cite{bai2025task}     & \cellcolor{myLightBlue}64.2353 & 3.8827 & 6.7790 & \cellcolor{myLightRed}\textbf{0.4444} & 0.5133 & 0.0217 & \cellcolor{myLightRed}\textbf{62.0018} & 4.2478 & 6.8587 & 0.2898 & 0.2752 & 0.0122 \\
SAGE \cite{wu2025every}         & 63.0311 & 3,4917 & 6.2922 & 0.3606 & 0.4587 & 0.0194 & 60.2943 & 3.7956 & 6.8287 & 0.2645 & 0.2199 & 0.0117 \\ 
URFusion \cite{xu2025urfusion}  & 64.1650 & \cellcolor{myLightBlue}4.5382 & 6.7019 & 0.3810 & 0.4856 & \cellcolor{myLightBlue}0.0173 & 61.2389 & \cellcolor{myLightBlue}4.7825 & \cellcolor{myLightBlue}6.8965 & 0.2616 & \cellcolor{myLightBlue}0.3404 & \cellcolor{myLightBlue}0.0094 \\
OmniFuse \cite{zhang2025omnifuse} & 63.5446 & 3.1962 & 6.9400 & 0.3015 & 0.3811 & 0.0198 & 59.9495 & 3.3314 & 6.5837 & 0.1858 & 0.1849 & 0.1112 \\
ControlFusion \cite{tang2025controlfusion}& 63.8074 & 3.8100 & 6.5640 & 0.4100 & \cellcolor{myLightBlue}0.5385 & 0.0218 & 60.2807 & 4.1221 & 6.6777 & 0.2569 & 0.2656 & 0.0101 \\
\midrule
\textbf{Ours} & \cellcolor{myLightRed}\textbf{64.2640} & \cellcolor{myLightRed}\textbf{5.3664} & \cellcolor{myLightRed}\textbf{7.0657} & \cellcolor{myLightBlue}0.4181 & 0.4994 & \cellcolor{myLightRed}\textbf{0.0104} & \cellcolor{myLightBlue}61.2501 & \cellcolor{myLightRed}\textbf{5.8770} & \cellcolor{myLightRed}\textbf{7.1544} & \cellcolor{myLightRed}\textbf{0.3368} & \cellcolor{myLightRed}\textbf{0.4219} & \cellcolor{myLightRed}\textbf{0.0081} \\
\bottomrule
\end{tabular}
\end{table*}

\subsection{Standard Comparison}
To provide a comprehensive and objective evaluation of the proposed AD-SRF, we first conduct standard comparisons on the FMB and MFNet datasets. Given that the input data remains consistently degraded throughout, this set of experiments aims to validate the superiority of AD-SRF in both visual and semantic dimensions compared to the standard models of other methods. This comparison clearly demonstrates the baseline capabilities of various models when processing low-quality sources.

\begin{figure*}[h]
	\centering
	\includegraphics[width=1\linewidth]{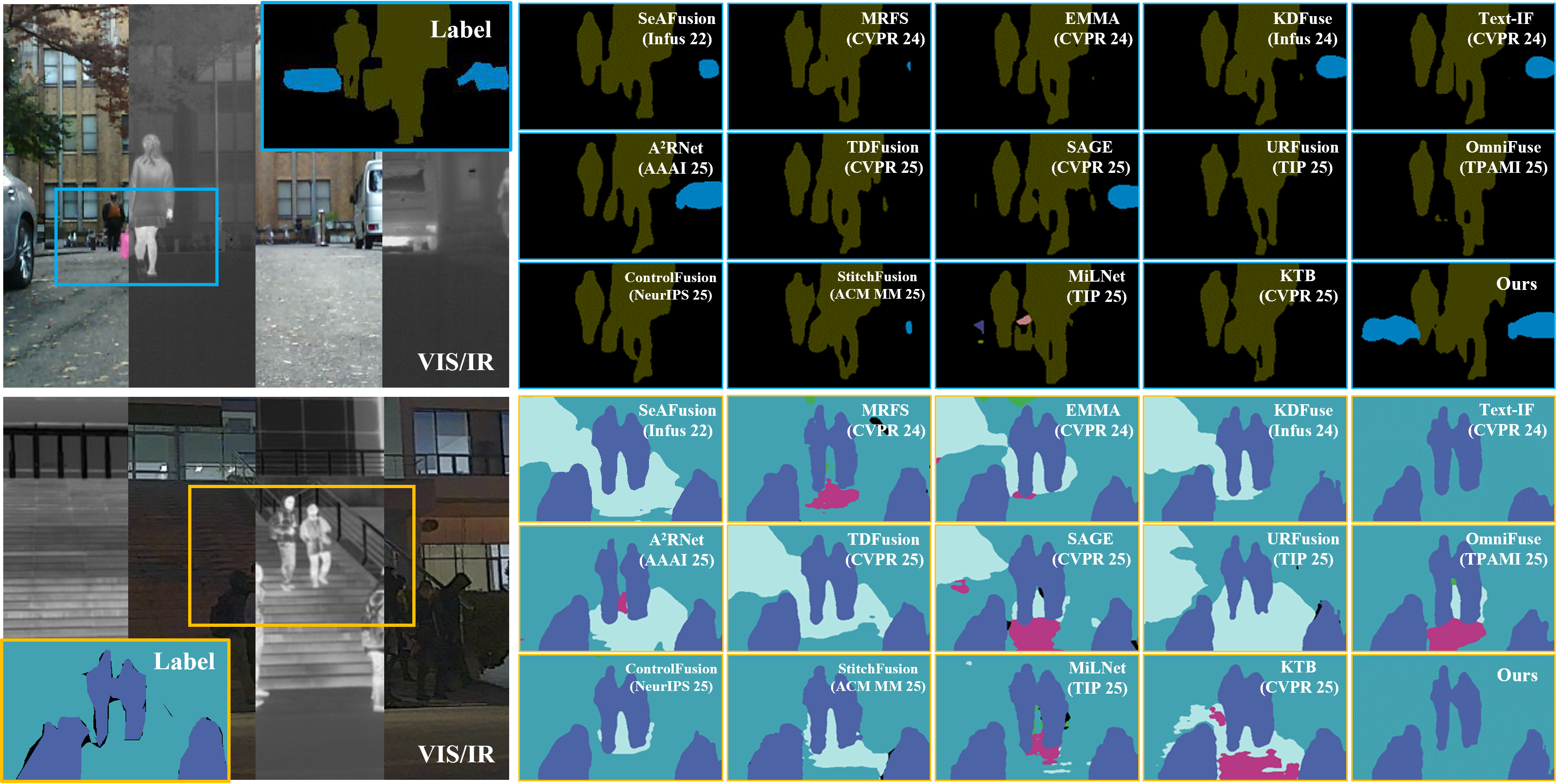}
	\vspace{-0.2in}
	\caption{Qualitative comparison of our method and competitive methods on semantic segmentation under standard setting.}\label{fig:5}
	\vspace{-2pt}
\end{figure*}

\begin{table}[h]
\vspace{-5pt}
\centering
\caption{Quantitative segmentation results on MFNet dataset under standard setting (mIoU is averaged over all 9 categories).}
\label{tab:MFNetseg}
\vspace{-5pt}
\renewcommand\theadgape{} 
\renewcommand\theadfont{\bfseries}
\renewcommand{\arraystretch}{1.1}
\setlength{\tabcolsep}{3pt} 
\resizebox{\linewidth}{!}{
\begin{tabular}{l|cccccccc|c} 
\toprule
\textbf{Methods} & \thead{Car} & \thead{Pers.} & \thead{Bike} & \thead{Curve} & \thead{C.S.} & \thead{Guar.} & \thead{Cone} & \thead{Bump} & \thead{mIoU} \\ 
\midrule
SeAFusion\cite{tang2022image} & 89.7 & 74.1 & 64.5 & 46.4 & 37.6 & 4.7 & 51.5 & 51.4 & 57.6 \\
MRFS \cite{zhang2024mrfs}      & 89.4 & 75.4 & 65.1 & 49.0 & 37.2 & 5.4 & 53.1 & 58.8 & 59.1 \\
EMMA \cite{zhao2024equivariant} & 89.2 & 72.6 & 63.4 & 47.6 & 29.4 & 6.3 & 54.6 & 59.8 & 57.9 \\
KDFuse \cite{yang2025kdfuse}   & 90.1 & 73.9 & 65.8 & 48.3 & 27.2 & 4.7 & 59.0 & 56.2 & 58.2 \\
Text-IF \cite{yi2024text}      & 90.1 & 72.9 & 65.2 & 47.2 & 34.7 & 5.4 & 55.6 & 54.1 & 58.2 \\
A$^2$RNet \cite{li2025a2rnet} & 90.1 & 72.9 & 65.2 & 47.2 & 34.7 & 5.4 & 55.6 & 54.1 & 58.2 \\
TDFusion \cite{bai2025task}    & 88.4 & 72.6 & 62.1 & 47.6 & 24.0 & 2.5 & 51.0 & 56.0 & 55.8 \\
SAGE \cite{wu2025every}        & 88.4 & 71.8 & 64.8 & 41.5 & 47.9 & 0.0 & 55.3 & 51.7 & 57.7 \\ 
URFusion \cite{xu2025urfusion} & 88.9 & 71.4 & 62.4 & 46.5 & 38.4 & 10.9 & 52.5 & 52.9 & 58.0 \\
OmniFuse \cite{zhang2025omnifuse} & 89.1 & 69.0 & 63.7 & 41.1 & 32.8 & 9.7 & 53.4 & 53.0 & 56.7 \\
ControlFusion \cite{tang2025controlfusion} & 89.6 & 72.5 & 67.4 & 48.1 & 37.2 & 9.7 & 57.4 & 54.2 & 59.4 \\
StitchFusion \cite{li2025stitchfusion}  & 88.9 & 74.0 & 63.3 & 45.1 & 36.1 & 0.4 & 55.6 & 55.2 & 57.4  \\
\rowcolor{myLightBlue}
MiLNet \cite{liu2025milnet}  & 87.9 & 72.3 & 64.5 & 46.2 & 45.8 & 10.1 & 55.8 & 60.1 & \underline{60.1} \\
KTB \cite{cai2025keep} & 90.6 & 74.3 & 66.3 & 51.2 & 39.4 & 8.9 & 53.7 & 56.3 & 59.9 \\
\midrule
\rowcolor{myLightRed}
\textbf{Ours} & 89.9 & 73.8 & 65.8 & 45.3 & 49.1 & 16.0 & 55.3 & 61.3 & \textbf{61.6} \\
\bottomrule
\end{tabular}
}
\vspace{-0.1in}
\end{table}

\begin{table*}[t] 
\vspace{-9pt}
\centering
\caption{Quantitative segmentation results on FMB dataset under standard setting (mIoU is averaged over all 15 categories). \\ Test set lacks bike class.}
\label{tab:FMBseg}
\vspace{-5pt}
\renewcommand\theadgape{} 
\renewcommand\theadfont{\bfseries}
\renewcommand{\arraystretch}{0.9}
\setlength{\tabcolsep}{4pt} 
\begin{tabular}{l|ccccccccccccccc|c} 
\toprule
\textbf{Methods} & \thead{BG.} & \thead{Road} & \thead{Side.} & \thead{Build.} & \thead{Lamp} & \thead{Sign} & \thead{Veg.} & \thead{Sky} & \thead{Pers.} & \thead{Car} & \thead{Truck} & \thead{Bus} & \thead{Motor.} & \thead{Bike} & \thead{Pole} & \thead{mIoU} \\ 
\midrule
SeAFusion\cite{tang2022image} & 15.75 & 86.71 & 52.08 & 83.53 & 49.29 & 75.23 & 86.30 & 94.61 & 69.38 & 83.56 & 54.59 & 74.40 & 39.79 & - & 50.93 & 61.08 \\
MRFS \cite{zhang2024mrfs}      & 17.04 & 86.94 & 46.28 & 85.39 & 50.07 & 75.75 & 86.98 & 95.11 & 71.31 & 76.16 & 34.44 & 28.63 & 49.03 & - & 53.64 & 57.12 \\
EMMA \cite{zhao2024equivariant} & 16.77 & 87.67 & 49.80 & 83.30 & 47.99 & 75.61 & 86.62 & 94.04 & 67.94 & 83.35 & 57.49 & 75.08 & 45.09 & - & 50.99 & 61.45\\
KDFuse \cite{yang2025kdfuse}   & 17.16 & 88.19 & 49.17 & 83.79 & 45.09 & 76.70 & 86.87 & 94.32 & 69.67 & 83.41 & 56.71 & 74.24 & 47.73 & - & 54.33 & 61.82 \\
Text-IF \cite{yi2024text}      & 16.37 & 87.65 & 49.99 & 85.17 & 46.48 & 75.37 & 85.75 & 94.08 & 68.94 & 83.48 & 57.97 & 74.90 & 48.07 & - & 54.25 & 61.90 \\
A$^2$RNet \cite{li2025a2rnet} & 18.42 & 88.02 & 49.68 & 83.29 & 40.77 & 77.06 & 86.55 & 93.85 & 68.38 & 82.37 & 43.37 & 73.05 & 41.03 & - & 52.59 & 59.89 \\
\rowcolor{myLightBlue}
TDFusion \cite{bai2025task}    & 16.08 & 88.01 & 53.27 & 84.36 & 46.89 & 76.25 & 86.66 & 94.04 & 69.23 & 83.94 & 57.61 & 77.92 & 44.61 & - & 53.30 & \underline{62.14} \\
SAGE \cite{wu2025every}        & 19.81 & 87.30 & 51.78 & 84.67 & 42.68 & 77.36 & 86.46 & 94.53 & 70.51 & 82.95 & 50.08 & 68.86 & 49.55 & - & 54.22 & 61.39 \\ 
URFusion \cite{xu2025urfusion} & 18.33 & 86.51 & 50.99 & 84.23 & 47.82 & 77.10 & 86.74 & 94.50 & 69.65 & 83.71 & 56.62 & 74.19 & 46.64 & - & 54.06 & 62.08 \\
OmniFuse \cite{zhang2025omnifuse} & 17.25 & 87.32 & 47.45 & 82.12 & 39.16 & 76.54 & 86.06 & 93.45 & 68.00 & 83.23 & 55.14 & 66.62 & 45.90 & - & 52.52 & 60.05 \\
ControlFusion \cite{tang2025controlfusion}& 17.23 & 88.52 & 49.54 & 84.34 & 48.79 & 76.57 & 86.27 & 94.16 & 71.00 & 84.20 & 55.48 & 71.09 & 45.94 & - & 52.01 & 61.67 \\
StitchFusion \cite{li2025stitchfusion}  & 15.38 & 89.43 & 52.68 & 83.89 & 49.69 & 74.67 & 87.12 & 94.98 & 72.94 & 82.63 & 45.43 & 34.34 & 50.72 & - & 53.04 & 59.13 \\
MiLNet \cite{liu2025milnet}  & 17.92 & 89.62 & 51.49 & 84.82 & 53.41 & 74.90 & 87.46 & 94.99 & 71.73 & 81.20 & 33.63 & 36.82 & 52.19 & - & 51.98  & 58.81 \\
KTB \cite{cai2025keep} & 16.43 & 89.76 & 51.53 & 85.99 & 43.25 & 75.54 & 87.57 & 95.19 & 69.55 & 84.75 & 50.18 & 79.45 & 44.96 & - & 51.29 & 61.73 \\
\midrule
\rowcolor{myLightRed}
\textbf{Ours} & 18.30 & 90.14 & 57.36 & 85.44 & 48.34 & 77.75 & 87.81 & 94.92 & 71.73 & 86.15 & 58.52 & 77.22 & 50.54 & - & 53.79 & \textbf{63.87} \\
\bottomrule
\end{tabular}
\end{table*}

\begin{table}[ht]
\vspace{3pt}
\centering
\caption{Quantitative results on FMB dataset under retraining enhancement setting.}
\label{tab:MetricFMBretrain}
\vspace{-3pt}

\resizebox{\columnwidth}{!}{
\renewcommand{\arraystretch}{1.2} 
\setlength{\tabcolsep}{4.5pt} 

\begin{tabular}{l|cccccc} 
\toprule
\multirow{2}{*}[-3pt]{\textbf{Methods}} & \multicolumn{6}{c}{\textbf{FMB}} \\
\cline{2-7}
& \rule{0pt}{11pt} \textbf{PSNR} $\uparrow$ & \textbf{AG} $\uparrow$  & \textbf{EN} $\uparrow$  & \textbf{SSIM} $\uparrow$  & $\bm{Q^{AB/F}}$ $\uparrow$  & $\bm{N^{AB/F}}$ $\downarrow$  \\ 

\midrule 

SeAFusion$^{\dagger}$\cite{tang2022image}    & 57.1016 & 5.3778 & 6.5741 & 0.2784 & 0.3469 & \cellcolor{myLightRed}\textbf{0.0073} \\
MRFS$^{\dagger}$ \cite{zhang2024mrfs}        & 60.4472 & 3.5689 & 6.8023 & 0.2535 & 0.2128 & 0.0122 \\
KDFuse$^{\dagger}$ \cite{yang2025kdfuse}     & 59.9691 & 4.9620 & 6.6208 & 0.2917 & 0.3674 & 0.0090 \\
Text-IF$^{\dagger}$ \cite{yi2024text}        & 60.5321 & 4.7523 & 7.1092 & 0.3117 & 0.3147 & 0.0080 \\
A$^2$RNet$^{\dagger}$ \cite{li2025a2rnet}       & \cellcolor{myLightRed}\textbf{62.0232} & 5.6666 & 6.9115 & 0.2325 & 0.3742 & 0.0096 \\
TDFusion$^{\dagger}$ \cite{bai2025task}      & 61.8351 & 4.9192 & \cellcolor{myLightRed}\textbf{7.2429} & 0.3160 & 0.3201 & 0.0084 \\
SAGE$^{\dagger}$ \cite{wu2025every}          & 61.6187 & 5.4439 & 6.7124 & 0.3135 & 0.4166 & 0.0115 \\ 
URFusion$^{\dagger}$ \cite{xu2025urfusion}   & 61.2457 & 4.4426 & 6.8938 & 0.2501 & 0.2973 & 0.0099 \\
OmniFuse$^{\dagger}$ \cite{zhang2025omnifuse} & \cellcolor{myLightBlue}62.0223 & 4.2787 & 6.7644 & 0.2752 & 0.2801 & 0.0120 \\
ControlFusion$^{\dagger}$ \cite{tang2025controlfusion}& 59.7308 & \cellcolor{myLightBlue}5.8443 & 6.9752 & \cellcolor{myLightBlue}0.3189 & \cellcolor{myLightRed}\textbf{0.5023} & 0.0078 \\
\midrule
\textbf{Ours} & 61.2501 & \cellcolor{myLightRed}\textbf{5.8770} & \cellcolor{myLightBlue}7.1544 & \cellcolor{myLightRed}\textbf{0.3368} & \cellcolor{myLightBlue}0.4219 & 0.0081 \\
\bottomrule
\end{tabular}
}
\vspace{-15pt}
\end{table}

\begin{table*}[t] 
\vspace{-6pt}
\centering
\caption{Quantitative segmentation results on FMB dataset under retraining enhancement setting (mIoU is averaged over all 15 categories). Test set lacks bike class.}
\label{tab:FMBsegretrain}
\vspace{-0.1in}
\renewcommand\theadgape{} 
\renewcommand\theadfont{\bfseries}
\renewcommand{\arraystretch}{1.1}
\setlength{\tabcolsep}{4pt} 
\begin{tabular}{l|ccccccccccccccc|c} 
\toprule
\textbf{Methods} & \thead{BG.} & \thead{Road} & \thead{Side.} & \thead{Build.} & \thead{Lamp} & \thead{Sign} & \thead{Veg.} & \thead{Sky} & \thead{Pers.} & \thead{Car} & \thead{Truck} & \thead{Bus} & \thead{Motor.} & \thead{Bike} & \thead{Pole} & \thead{mIoU} \\ 
\midrule
SeAFusion$^{\dagger}$\cite{tang2022image}      & 17.50 & 88.83 & 52.80 & 84.07 & 49.22 & 76.46 & 87.07 & 94.36 & 69.12 & 82.20 & 55.23 & 59.43 & 42.49 & - & 55.03 & 60.92 \\
MRFS$^{\dagger}$ \cite{zhang2024mrfs}          & 16.99 & 86.84 & 46.15 & 85.35 & 48.83 & 75.85 & 86.88 & 95.11 & 71.18 & 76.18 & 34.75 & 28.52 & 48.50 & - & 53.35 & 56.96 \\
KDFuse$^{\dagger}$ \cite{yang2025kdfuse}       & 17.87 & 88.05 & 52.17 & 82.51 & 46.84 & 76.50 & 86.47 & 94.14 & 68.76 & 83.24 & 58.57 & 61.98 & 44.94 & - & 53.01 & 61.00 \\
Text-IF$^{\dagger}$ \cite{yi2024text}          & 17.09 & 88.03 & 49.92 & 83.85 & 48.86 & 75.65 & 86.97 & 94.08 & 70.66 & 83.19 & 57.40 & 68.84 & 42.57 & - & 54.80 & 61.46 \\
A$^2$RNet$^{\dagger}$ \cite{li2025a2rnet}   & 15.85 & 85.49 & 48.44 & 82.53 & 40.66 & 74.88 & 85.86 & 94.47 & 66.11 & 81.83 & 59.47 & 65.29 & 47.12 & - & 52.68 & 60.04 \\
\rowcolor{myLightBlue}
TDFusion$^{\dagger}$ \cite{bai2025task}        & 17.02 & 88.22 & 51.38 & 85.21 & 47.38 & 75.05 & 87.07 & 94.37 & 69.70 & 83.98 & 58.87 & 76.50 & 44.72 & - & 53.19 & \underline{62.18} \\
SAGE$^{\dagger}$ \cite{wu2025every}            & 18.19 & 87.17 & 50.93 & 83.89 & 48.96 & 75.94 & 86.76 & 94.35 & 69.02 & 83.56 & 53.42 & 75.50 & 49.49 & - & 53.15 & 62.02 \\ 
URFusion$^{\dagger}$ \cite{xu2025urfusion}     & 16.97 & 86.99 & 51.88 & 84.95 & 46.71 & 76.59 & 86.65 & 94.09 & 68.64 & 82.62 & 54.62 & 76.19 & 43.02 & - & 53.97 & 61.73 \\
OmniFuse$^{\dagger}$ \cite{zhang2025omnifuse}  & 17.29 & 88.35 & 49.11 & 84.23 & 45.75 & 76.96 & 86.51 & 94.01 & 68.74 & 83.95 & 56.59 & 75.98 & 44.81 & - & 54.79 & 61.80 \\
ControlFusion$^{\dagger}$ \cite{tang2025controlfusion}& 17.24 & 88.95 & 52.88 & 84.90 & 47.55 & 76.24 & 87.62 & 94.51 & 70.29 & 83.47 & 50.80 & 70.31 & 44.04 & - & 52.38 & 61.41 \\
\midrule
\rowcolor{myLightRed}
\textbf{Ours} & 18.30 & 90.14 & 57.36 & 85.44 & 48.34 & 77.75 & 87.81 & 94.92 & 71.73 & 86.15 & 58.52 & 77.22 & 50.54 & - & 53.79 & \textbf{63.87} \\
\bottomrule
\end{tabular}
\end{table*}

 \textbf{Visual Comparison.} We first conduct visual comparative experiments against eleven state-of-the-art multimodal image fusion algorithms. On MFNet dataset, we focus on challenging low-light scenarios that typically compromise comprehensive scene representation. As shown in Fig. \ref{fig:4}, for MFNet dataset, although recent methods like MRFS, URFusion, OmniFuse and ControlFusion achieve a certain degree of restoration, our approach performs significantly better in preserving and boosting subtle textures and structural components within dark regions, as well as maintaining superior color fidelity. Similarly, for FMB dataset, we evaluate performance under foggy conditions, as shown in Fig. \ref{fig:4}. Benefiting from the comprehensive constraints of $\mathcal{L}_{f}$ across three dimensions, our method demonstrates a clear advantage in local contrast enhancement and background reconstruction. 
Moreover, the  quantitative comparison in Table \ref{tab:MetricComp} shows that our AD-SRF achieves the top rank in 4 and 5 out of the 6 evaluated metrics on  MFNet and FMB datasets, respectively. These superior results underscore the effectiveness of our RVF network. In summary, the consistent performance across diverse datasets validates the superiority of our unified restoration-fusion design. By co-optimizing image restoration and fusion through the collaborative loss $\mathcal{L}_{f}$, our framework achieves a dual utility: it recovers a visually natural appearance for human observation while simultaneously constructing a solid foundation that facilitates robust downstream interpretation.

\textbf{Semantic Evaluation. } 
We then evaluate AD-SRF's semantic performance against eleven fusion-based and three RGB-T segmentation methods. Benefiting from the ADG mechanism's capability for semantic mining and feature replenishment, our method exhibits a huge advantage in complex scenes. As shown in Fig. \ref{fig:5}, our approach stands out as the only method capable of accurately detecting the bicycle to the left of the pedestrian in MFNet dataset, even when faced with low-resolution inputs and significant intrinsic degradation. Furthermore, in FMB low-light scenarios, our framework precisely delineates boundaries, eliminating common errors on facades.  This is primarily because the antagonistic interplay within the ADG mechanism forces the model to proactively mine semantic vulnerabilities and replenish task-relevant features that are typically discarded during the fusion process.
Quantitatively, as presented in Table \ref{tab:MFNetseg} and \ref{tab:FMBseg}, our approach surpasses the runner-up by 1.5 and 1.73 mIoU scores on MFNet and FMB datasets, respectively. Notably, A$^2$RNet focuses on enhancing robustness against external perturbations but fails to effectively improve inherent semantic attributes, resulting in inferior segmentation performance compared to AD-SRF. All these results substantiate that our AD-SRF does not merely improve visual appearance but fundamentally elevates the semantic performance of the fused representation. By bridging the gap between low-level vision and high-level semantic requirements through adversarial training, our method establishes a new performance 
ceiling for multimodal perception tasks.

\subsection{Retraining Enhancement Comparison}
To further validate the superiority of AD-SRF, we extended our evaluation beyond a direct comparison with the original results of competing algorithms on FMB dataset. Moreover, we retrained 10 representative fusion methods using the high-quality images provided in this study. Notably, among the 11 methods previously mentioned, EMMA was excluded from this retraining phase because its architecture lacks constraints tied directly to the source images. This evaluation setup not only mitigates the impact of data bias but also unveils the intrinsic potential of different fusion architectures when handling degraded inputs. The entire evaluation phase is conducted from both visual and semantic perspectives. Specifically, we utilize high-quality images as ground-truth labels to guide the retraining of each algorithm. Subsequently, the semantic segmentation network is retrained using the generated fused images as input to evaluate their semantic utility.

\begin{figure*}[t]
	\vspace{-0.06in}
	\centering
	\includegraphics[width=0.99\linewidth]{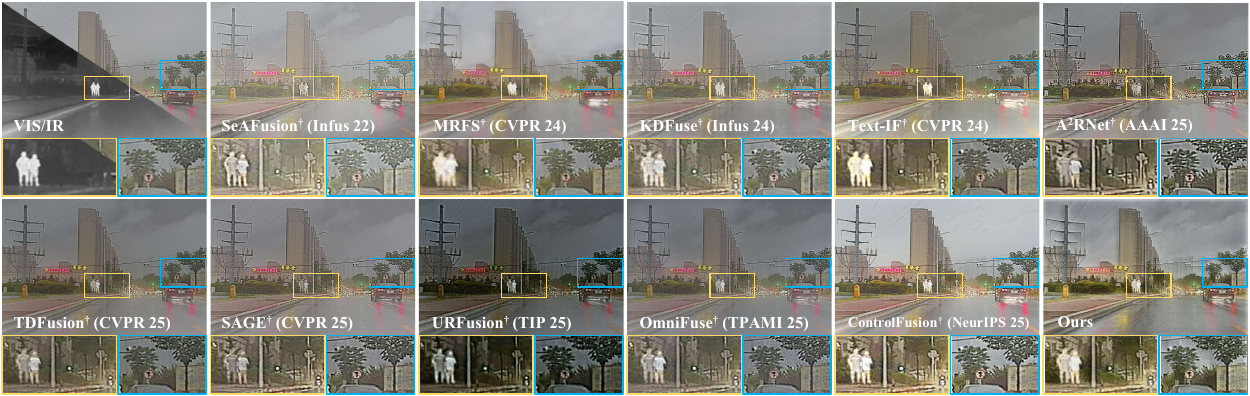}
	\vspace{-0.07in}
	\caption{Qualitative comparison of our method and existing image fusion methods under retraining enhancement setting.}\label{fig:6}
\end{figure*}

\textbf{Visual Comparison.} Qualitatively, as illustrated in Fig. \ref{fig:6}, in contrast to other methods, AD-SRF exhibits a superior capacity to proactively reconstruct sharper geometric boundaries and more natural chromatic distributions from degraded inputs. This advantage is especially pronounced in the informative richness of the background details shown in the enlarged regions. Furthermore, as detailed in Table \ref{tab:MetricFMBretrain}, benefiting from the cross-modal retrieval design of the RVF baseline and the guidance of the restoration-fusion collaborative loss, our method achieves outstanding performance, ranking first in two and second in another two out of the six objective metrics, further validating the superiority of its visual quality.

\begin{figure*}[!t]
	\vspace{-0.09in}
	\centering
	\includegraphics[width=1\linewidth]{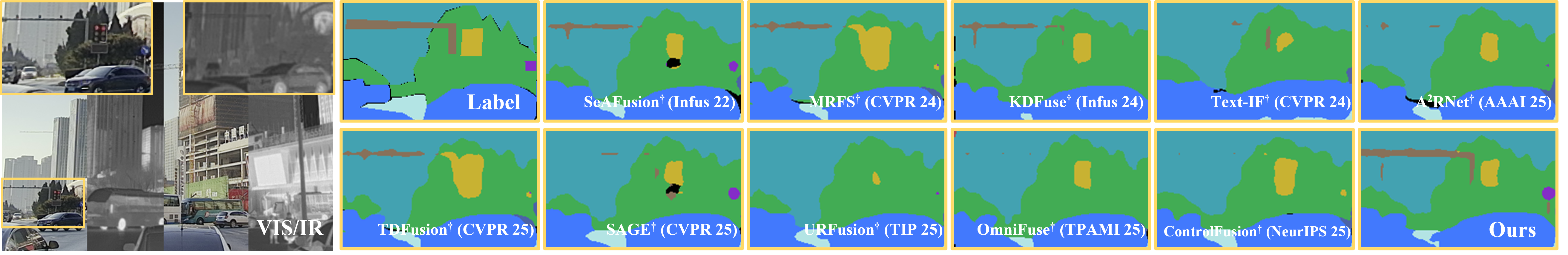}
	\vspace{-0.23in}
	\caption{Qualitative comparison of our method and existing image fusion methods under retraining enhancement setting.}\label{fig:7}
	\vspace{-10pt}
\end{figure*}

\textbf{Semantic Evaluation.} Attributed to the semantic enhancement derived from the attack-defense interplay, our approach qualitatively stands out as the sole method able to completely identify detailed objects including poles and signs in enlarged regions, as illustrated in Fig. \ref{fig:7}. Furthermore, it quantitatively outperforms the second-best performing method by 1.67 in mIoU score, as shown in Table \ref{tab:FMBsegretrain}. These results directly validate the effectiveness and necessity of the ADG mechanism within the AD-SRF framework for semantic reinforcement.

\subsection{Adversarial Semantic Validation} 
\begin{figure*}[t]
	\centering
	\includegraphics[width=1\linewidth]{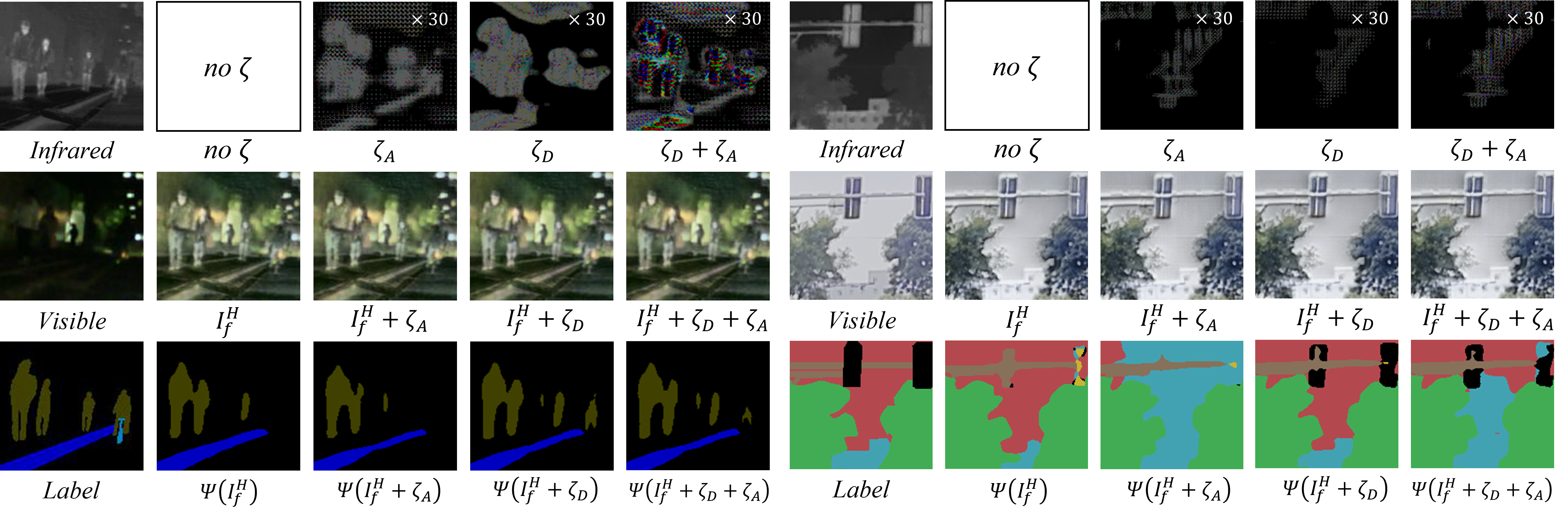}
	\vspace{-0.28in}
	\caption{Qualitative validation of ADG paradigm: visual fidelity and semantic interaction.}\label{fig:8}
\end{figure*}

This section aims to investigate the impact of the perturbations produced during ADG training on both final semantic performance and visual effects.
Fig. \ref{fig:8} presents the visual and semantic visualization results of our method. The comparison between the visualized $\zeta$ (Row 1) and the perturbation-added images (Row 2) against the clean counterparts demonstrates that the generated $\zeta$ is of minimal magnitude, rendering its impact on image quality visually imperceptible. Moreover, the segmentation visualizations in Row 3 verify the effectiveness of the generated $\zeta$ in modulating semantic attributes. Simultaneously, we performed a quantitative assessment by calculating the SSIM and PSNR between the perturbation-added images and the original ones. As reported in Table \ref{tab:Visual_Fidelity_Eval}, the results further corroborate that the $\zeta$ derived via the ADG mechanism exerts a negligible influence on visual fidelity.  Finally, the metrics in Table \ref{tab:SemanticADG} validate the distinct functional roles of the learned perturbations: the destructive capability of $\zeta_A$ and the collaborative reinforcement-defense synergy of $\zeta_D$. Such findings validate that AD-SRF can refine semantic representations while preserving visual fidelity, establishing a vital foundation for attaining the dual-objective of superior visual quality and robust semantic decision.

\begin{table}[t]
\vspace{-15pt}
\centering
\caption{Visual fidelity validation of ADG paradigm.}\label{tab:Visual_Fidelity_Eval}
\vspace{-2pt}
\renewcommand{\arraystretch}{0.8} 
\setlength{\tabcolsep}{8pt}
\begin{tabular}{l|cc|cc} 
\toprule
\multirow{2}{*}[-3pt]{\textbf{Methods}} & \multicolumn{2}{c|}{\textbf{MFNet}} & \multicolumn{2}{c}{\textbf{FMB}} \\
\cline{2-5}
& \rule{0pt}{11pt} \textbf{PSNR} $\uparrow$ & \textbf{SSIM} $\uparrow$ & \textbf{PSNR} $\uparrow$ & \textbf{SSIM} $\uparrow$ \\
\midrule
$I_f^H {+} \zeta_A$               & \textbf{48.72} & \textbf{0.9949} & \textbf{51.34} & \textbf{0.9948} \\
$I_f^H {+} \zeta_D$               & \underline{47.02} & \underline{0.9910} & \underline{50.96} & \underline{0.9942} \\
$I_f^H {+} \zeta_D {+} \zeta_A$    & 45.64 & 0.9883 & 49.31 & 0.9913 \\
\bottomrule 
\end{tabular}
\end{table}

\begin{table}[t]
\vspace{-6pt}
\centering
\caption{Semantic interaction validation of ADG paradigm.}
\label{tab:SemanticADG}
\vspace{-2pt}
\footnotesize
\renewcommand{\arraystretch}{0.8}
\setlength{\tabcolsep}{4pt}       
\begin{tabular}{l|cccc}
\toprule
\textbf{Datasets} & $\Psi(I_f^H)$ & $\Psi(I_f^H {+} \zeta_A)$ & $\Psi(I_f^H {+} \zeta_D)$ & $\Psi(I_f^H {+} \zeta_D {+} \zeta_A)$ \\
\midrule
\textbf{MFNet} & 59.61 & 57.77 & \textbf{61.65} & \underline{61.01} \\
\textbf{FMB}   & 62.27 & 60.39 & \textbf{63.87} & \underline{63.72} \\
\bottomrule
\end{tabular}
\end{table}

\subsection{Application to Other Fusion Tasks}
To further validate the generalizability of the AD-SRF framework, we extended our evaluation to two representative and challenging domains: Polarization Image Fusion and Near-Infrared/Visible (NIR-VIS) Image Fusion. These zero-shot experiments are designed to demonstrate that our RVF baseline can consistently produce high-quality images with significant visual advantages by synergistically rectifying sensor degradation and precisely integrating cross-modal features.

\begin{figure*}[t]
	\centering
	\includegraphics[width=1\linewidth]{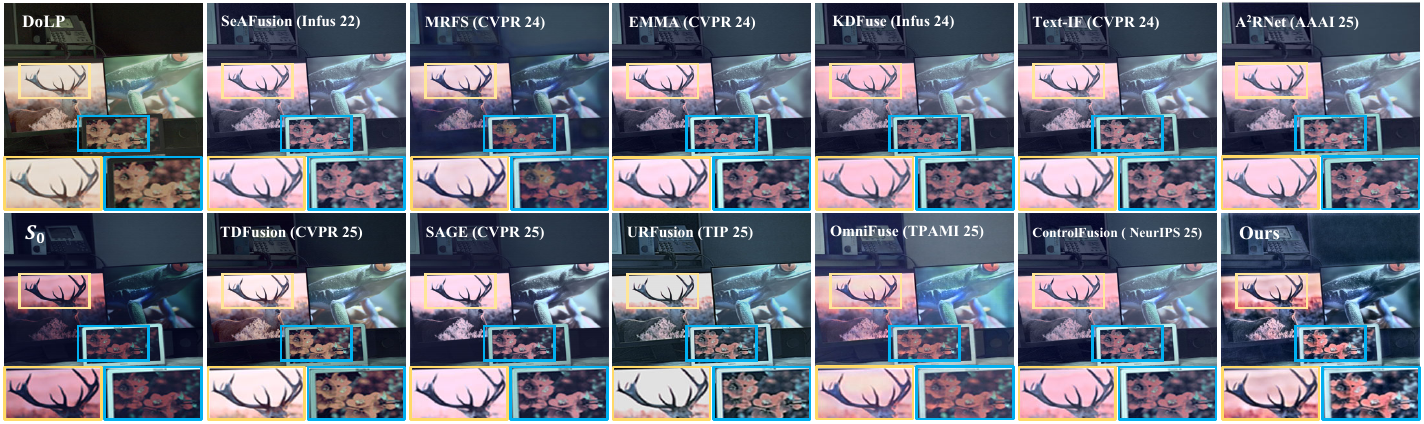}
	\vspace{-0.2in}
	\caption{Qualitative applications to the task of polarization image fusion.}\label{fig:9}
\end{figure*}

\textbf{Polarization Image Fusion.} Polarization fusion aims to leverage the physical priors in the Degree of Linear Polarization (DoLP) to compensate for the deficiencies of the intensity image ($S_0$) in detail and material discrimination. In contrast to existing fusion methods that often suffer from washed-out colors and diminished contrast due to inherent sensor degradation, the RVF baseline within AD-SRF exhibits superior restoration capabilities. A close inspection of Fig. \ref{fig:9} highlights this critical difference: while competitor approaches struggle to resolve the fine details and natural hues of complex objects, such as the flowers inside the frame, resulting in a flat and desaturated appearance, our method successfully reconstructs them with vibrant color and distinct contrast. By effectively transforming the intrinsic physical features of the DoLP into visually salient textures, AD-SRF delivers fused results with substantial quantitative advantages in information content and global contrast, as confirmed in Table \ref{tab:Metricpolar}.

\begin{figure*}[t]
	\centering
	\includegraphics[width=1\linewidth]{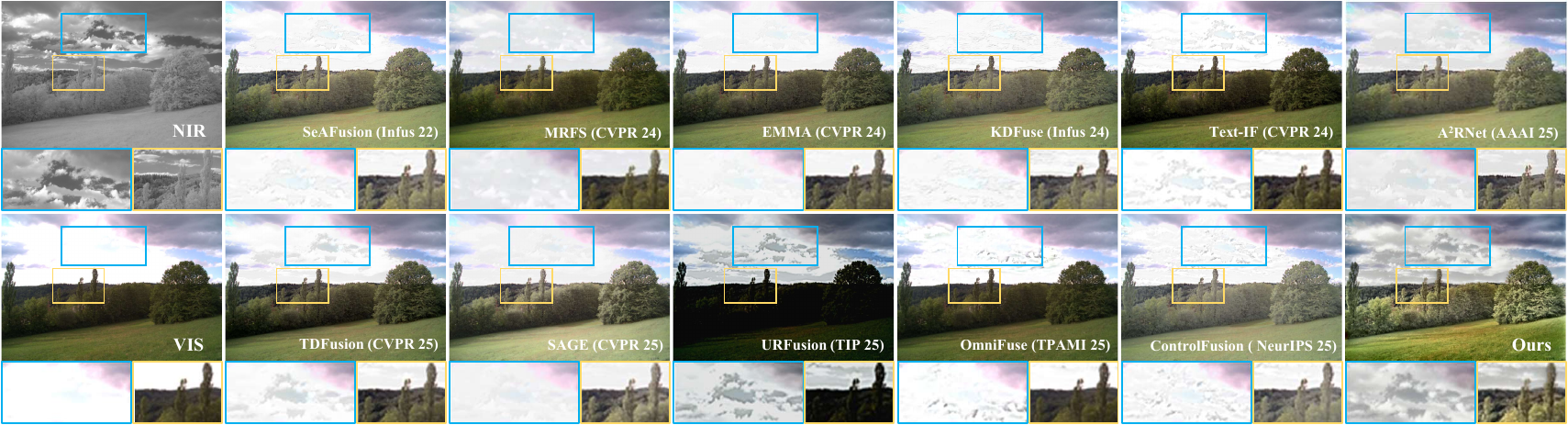}
	\vspace{-0.2in}
	\caption{Qualitative applications to the task of near-infrared and visible image fusion.}\label{fig:10}
\end{figure*}

\begin{table}[t]
\vspace{-15pt}
\centering
\caption{Quantitative results on Polarization image fusion.}
\label{tab:Metricpolar}
\vspace{-2pt}
\resizebox{\columnwidth}{!}{
\renewcommand{\arraystretch}{1.2} 
\setlength{\tabcolsep}{4.5pt} 
\begin{tabular}{l|cccccc} 
\toprule
\multirow{2}{*}[-3pt]{\textbf{Methods}} & \multicolumn{6}{c}{\textbf{Polarization}} \\
\cline{2-7}
& \rule{0pt}{11pt} \textbf{PSNR} $\uparrow$ & \textbf{AG} $\uparrow$  & \textbf{EN} $\uparrow$  & \textbf{SSIM} $\uparrow$  & $\bm{Q^{AB/F}}$ $\uparrow$  & $\bm{N^{AB/F}}$ $\downarrow$  \\ 
\midrule 
SeAFusion\cite{tang2022image}    & 62.0774 & 2.2667 & 6.5333 & 0.2711 & 0.2265 & 0.1661 \\
MRFS \cite{zhang2024mrfs}        & 61.5999 & 1.4195 & 6.7700 & 0.2224 & 0.1628 & \cellcolor{myLightBlue}0.1412 \\
EMMA \cite{zhao2024equivariant}  & 61.7656 & 2.1649 & 6.7912 & 0.2624 & 0.2210 & 0.1501 \\
KDFuse \cite{yang2025kdfuse}     & 62.2455 & 2.1002 & 6.5313 & 0.2589 & 0.2197 & 0.1738 \\
Text-IF \cite{yi2024text}        & 62.0376 & 2.4223 & 6.7537 & \cellcolor{myLightBlue}0.3094 & \cellcolor{myLightRed}\textbf{0.2423} & 0.1662 \\
A$^2$RNet \cite{li2025a2rnet}       & \cellcolor{myLightBlue}62.3309 & 1.4572 & 6.4252 & 0.1332 & 0.1423 & 0.1452 \\
TDFusion \cite{bai2025task}      & 61.9120 & \cellcolor{myLightBlue}2.4683 & \cellcolor{myLightBlue}6.8179 & \cellcolor{myLightRed}\textbf{0.3351} & \cellcolor{myLightBlue}0.2380 & 0.1545 \\
SAGE \cite{wu2025every}          & 61.7779 & 1.7519 & 6.0907 & 0.2532 & 0.1834 & 0.1457 \\ 
URFusion \cite{xu2025urfusion}   & 60.9140 & 1.8705 & 6.6363 & 0.2107 & 0.2021 & \cellcolor{myLightRed}\textbf{0.1405} \\
OmniFuse \cite{zhang2025omnifuse} & 61.7265 & 1.2942 & 6.7474 & 0.1268 & 0.1534 & 0.1487 \\
ControlFusion \cite{tang2025controlfusion}& 61.3981 & 2.1071 & 6.6644 & 0.2239 & 0.2191 & 0.1669 \\
\midrule
\textbf{Ours} & \cellcolor{myLightRed}\textbf{62.3544} & \cellcolor{myLightRed}\textbf{2.5759} & \cellcolor{myLightRed}\textbf{7.0030} & 0.2322 & 0.2183 & 0.1489 \\
\bottomrule
\end{tabular}
}
\end{table}

\begin{table}[t]
\centering
\caption{Quantitative results on NIR-VIS image fusion.}
\label{tab:MetricNIRVIS}
\vspace{-0.07in}

\resizebox{\columnwidth}{!}{
\renewcommand{\arraystretch}{1.3} 
\setlength{\tabcolsep}{4.5pt} 

\begin{tabular}{l|cccccc} 
\toprule
\multirow{2}{*}[-3pt]{\textbf{Methods}} & \multicolumn{6}{c}{\textbf{NIR-VIS}} \\
\cline{2-7}
& \rule{0pt}{11pt} \textbf{PSNR} $\uparrow$ & \textbf{AG} $\uparrow$  & \textbf{EN} $\uparrow$  & \textbf{SSIM} $\uparrow$  & $\bm{Q^{AB/F}}$ $\uparrow$  & $\bm{N^{AB/F}}$ $\downarrow$  \\ 
\midrule 
SeAFusion\cite{tang2022image}    & 65.1848 & 5.5793 & 7.2054 & 0.5399 & 0.3544 & \cellcolor{myLightBlue}0.0460 \\
MRFS \cite{zhang2024mrfs}        & 64.2899 & 3.9187 & 7.1277 & 0.4324 & 0.2219 & 0.0570 \\
EMMA \cite{zhao2024equivariant}  & 65.1122 & \cellcolor{myLightBlue}5.6443 & 7.2679 & 0.5326 & 0.3594 & 0.0479 \\
KDFuse \cite{yang2025kdfuse}     & 64.8639 & 4.9950 & 7.1430 & 0.4913 & 0.2981 & 0.0554 \\
Text-IF \cite{yi2024text}        & 64.7539 & 5.3033 & 7.2012 & 0.5240 & 0.3321 & 0.0495 \\
A$^2$RNet \cite{li2025a2rnet}       & \cellcolor{myLightBlue}64.8192 & 3.3725 & 7.0902 & 0.3430 & 0.1685 & 0.0562 \\
TDFusion \cite{bai2025task}      & \cellcolor{myLightRed}\textbf{65.5393} & 5.6095 & \cellcolor{myLightBlue}7.3234 & \cellcolor{myLightRed}\textbf{0.5758} & \cellcolor{myLightBlue}0.3716 & 0.0483 \\
SAGE \cite{wu2025every}          & \cellcolor{myLightBlue}65.2393 & 4.8510 & 7.2029 & 0.5093 & 0.3087 & 0.0543 \\ 
URFusion \cite{xu2025urfusion}   & 62.5243 & 4.3464 & 6.7862 & 0.3853 & 0.2376 & 0.0527 \\
OmniFuse \cite{zhang2025omnifuse} & 64.4191 & 3.6144 & 7.1191 & 0.3566 & 0.1958 & 0.0553 \\
ControlFusion \cite{tang2025controlfusion}& 64.4189 & 5.0829 & 7.1435 & 0.5037 & 0.3237 & 0.0534 \\
\midrule
\textbf{Ours} & 64.5457 & \cellcolor{myLightRed}\textbf{7.1130} & \cellcolor{myLightRed}\textbf{7.5281} & \cellcolor{myLightBlue}0.5612 & \cellcolor{myLightRed}\textbf{0.4236} & \cellcolor{myLightRed}\textbf{0.0385} \\
\bottomrule
\end{tabular}
}
\vspace{0.02in}
\end{table}

\textbf{NIR-VIS Image Fusion.} NIR-VIS fusion aims for spectral complementarity, utilizing the structural integrity of NIR to compensate for VIS degradation while preserving natural chromatic information to achieve all-weather perception. In contrast to existing methods that often struggle to balance illumination enhancement with detail preservation, the RVF baseline demonstrates superior performance in rectifying localized underexposure while integrating cross-modal features. As illustrated in Fig. \ref{fig:10}, while competing approaches often leave the underexposed grassy areas in deep shadow with indistinguishable details, our method exhibits robust illumination recovery, significantly enhancing visibility and restoring the natural appearance of the terrain. Simultaneously, owing to the collaborative cross-modal constraints, AD-SRF maximizes the preservation of intricate cloud textures from the NIR band. As reported in
Table \ref{tab:MetricNIRVIS}, our method achieves outstanding performance, ranking first in four and second in one out of the six quantitative metrics, which fully underscores the potent restoration-fusion synergy and robust generalization capability of the proposed RVF baseline.

\subsection{Ablation Studies} 
\begin{figure*}[t]
	\centering
	\includegraphics[width=1\linewidth]{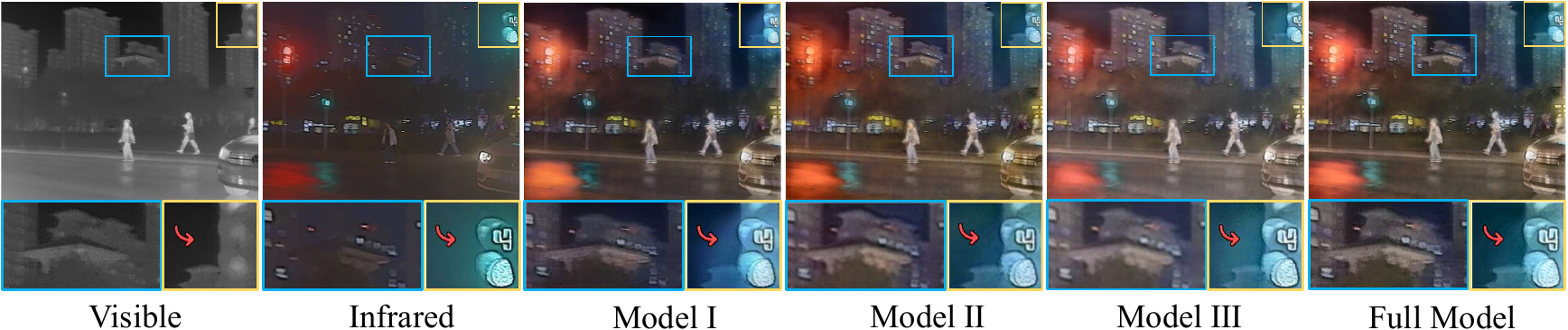}
	\caption{Qualitative results of ablations on visual components.}\label{fig:11}
	\vspace{-0.1in}
\end{figure*}

\textbf{Visual Ablation.}
To validate the efficacy and superiority of the constituent components within our proposed RVF baseline in preserving visual quality and facilitating image restoration, we evaluate the following model variants. \textbf{Model I:} Adopts the traditional coupled training paradigm, where the parameters of the fusion and segmentation networks are updated simultaneously during the training process. \textbf{Model II:} Replaces the proposed CMR module with a standard convolutional structure. \textbf{Model III:} Substitutes the proposed $\mathcal{L}_{con}$ with a common max-intensity loss, formulated as $\mathcal{L}_{max} = \| I_{f}^H - \max(I^{H}_{vis}, I^{H}_{ir}) \|_1$. The qualitative and quantitative results are summarized in Fig. \ref{fig:11} and Table \ref{tab:Ablation}. As observed, Model I prioritizes semantic performance enhancement at the expense of certain image restoration capabilities compared to the pure visual baseline. Model II exhibits inferior performance in architectural contrast within dark regions and the suppression of spurious artifacts near traffic lights compared to the Full Model, thereby validating the superior capability of the CMR module in extracting and integrating cross-modal features for the restoration-fusion task. Furthermore, $\mathcal{L}_{max}$ employed in Model III represents a crude pixel-level maximization approach, which leads to blurred details and severe ghosting artifacts, which are detrimental to human visual perception. Quantitatively, the Full Model achieves three first-place and three second-place rankings across the six metrics, significantly outperforming all other ablation variants. Collectively, these findings demonstrate that the proposed decoupled paradigm, structural design and constraint strategies are indispensable for maintaining high-fidelity visual quality.

\begin{figure}[t] 
	\centering
	\includegraphics[width=1\columnwidth]{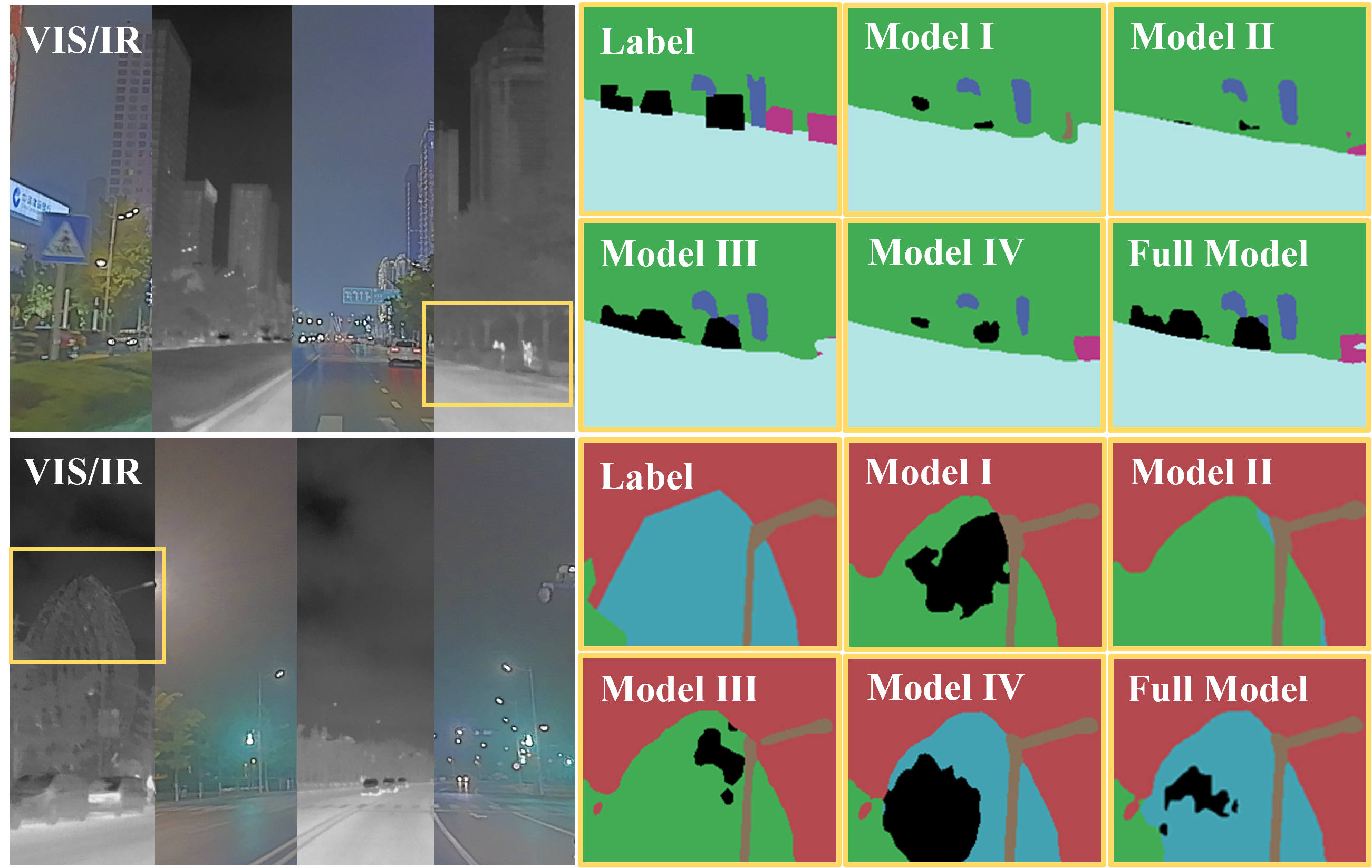} 
	\caption{Qualitative results of ablations on semantic components.}
	\label{fig:12}
\end{figure}

\begin{table}[!ht]
	\centering
	\caption{Quantitative results of ablations on visual components.}
	\label{tab:Ablation}
	\vspace{-0.05in}
	\renewcommand{\arraystretch}{1} 
	\resizebox{\columnwidth}{!}{
		\begin{tabular}{c|cccccc} 
			\toprule
			\textbf{Ablations} & \textbf{PSNR} $\uparrow$ & \textbf{AG} $\uparrow$ & \textbf{EN} $\uparrow$ & \textbf{SSIM} $\uparrow$ & $\bm{Q^{AB/F}}$ $\uparrow$ & $\bm{N^{AB/F}}$ $\downarrow$ \\ 
			\midrule 
			Model I    & \textbf{62.0865} & 5.4873 & \underline{7.0925} & 0.3293 & 0.4028 & 0.0093 \\
			Model II     & 61.1906 & 5.4129 & 6.9576 & 0.3257 & 0.3766 & \underline{0.0083} \\
			Model III    & 59.3522 & \textbf{5.9259} & 7.0201 & \underline{0.3359} & \textbf{0.4408} & 0.0087 \\
			Full Model  & \underline{61.2501} & \underline{5.8770} & \textbf{7.1544} & \textbf{0.3368} & \underline{0.4219} &  \textbf{0.0081} \\			
			\bottomrule
		\end{tabular}
	}
\end{table}

\begin{table}[!ht]
	\vspace{-5pt}
	\centering
	\caption{Quantitative results of ablations on semantic components.}
	\label{tab:Ablation_Semantic}
	\vspace{-0.05in}
	\footnotesize
	\renewcommand{\arraystretch}{1}
	\setlength{\tabcolsep}{2pt}
	\begin{tabular}{c|cccccccc|c}
		\toprule		
		\textbf{Ablations} & \textbf{Road} & \textbf{Side.} & \textbf{Build.} & \textbf{Sign} & \textbf{Veg.} & \textbf{Sky} & \textbf{Pers.} & \textbf{Car} & \textbf{mIoU} \\		
		\midrule		
		Model I  & 89.09 & 53.50 & 84.66 & 76.60 & 87.81 & 94.53 & 70.50 & 85.58 & 62.92 \\		
		Model II   & 87.85 & 48.26 & 85.01 & 77.39 & 87.10 & 94.54 & 69.91 & 83.28 & 62.27 \\		
		Model III & 90.51 & 55.61 & 84.79 & 76.40 & 87.05 & 94.86 & 69.99 & 85.86 & 63.00 \\		
		Model IV  & 89.57 & 56.80 & 85.11 & 77.59 & 87.44 & 94.79 & 71.51 & 85.82 & \underline{63.33} \\		
		Full Model & 90.14 & 57.36 & 85.44 & 77.75 & 87.81 & 94.92 & 71.73 & 86.15 & \textbf{63.87} \\		
		\bottomrule		
	\end{tabular}
\end{table}

\textbf{Semantic Ablation.} To thoroughly assess the contribution of the ADG mechanism to downstream segmentation tasks, we compare the following model variants. \textbf{Model I:} The same as the variant in visual ablation, adopts the conventional coupled training paradigm. \textbf{Model II:} Utilizes only the frozen visual baseline $\Phi$ from Phase I for feature extraction and directly trains the downstream segmentation network. \textbf{Model III:} Removes the ADG mechanism and trains the defender $\mathcal{D}$ from scratch as a semantic reinforcement module. \textbf{Model IV:} Fixes the parameters of a fully-capable pre-trained attacker $\mathcal{A}$ and trains the defender $\mathcal{D}$ against these static perturbations. The qualitative and quantitative results are summarized in Fig. \ref{fig:12} and Table \ref{tab:Ablation_Semantic}. The experimental findings reveal several key insights. First, although Model I attempts to optimize both tasks simultaneously, the performance gains remain marginal, likely due to optimization interference within the coupled architecture. Second, compared to the pure baseline (Model II), the Full Model achieves a substantial leap in mIoU (+1.60), revealing the indispensable necessity of semantic reinforcement. Furthermore, due to the absence of the attacker's guidance, the defender in Model III is unable to reach the model's performance bottlenecks, resulting in limited improvements. Finally, while Model IV introduces a fully-capable static attacker, it is still outperformed by the Full Model. This demonstrates the critical importance of dynamic alternating optimization within the ADG mechanism, which compels the network to capture robust, intrinsic features through co-evolutionary competition. Collectively, these results prove that each component of the proposed ADG mechanism significantly contributes to the enhancement of semantic performance.

\begin{table}[t]
	\vspace{-2pt}
	\centering
	\caption{Quantitative evaluation of  ADG's generalization performance across various fusion methods.}
	\label{tab:Methods_Comparison}
	\footnotesize
	\renewcommand{\arraystretch}{1}
	\setlength{\tabcolsep}{1.8pt}
	\begin{tabular}{l|cccccccc|c}
		\toprule
		\textbf{Methods} & \textbf{Road} & \textbf{Side.} & \textbf{Build.} & \textbf{Sign} & \textbf{Veg.} & \textbf{Sky} & \textbf{Pers.} & \textbf{Car} & \textbf{mIoU} \\
		\midrule		
		KDFuse        & 88.19 & 49.17 & 83.79 & 76.70 & 86.87 & 94.32 & 69.67 & 83.41 & 61.82 \\
		\rowcolor{gray!15} KDFuse$^{*}$  & 90.43 & 54.50 & 83.98 & 76.84 & 87.49 & 94.12 & 70.06 & 84.74 & 63.06 \\		
		Text-IF       & 87.65 & 49.99 & 85.17 & 75.37 & 85.75 & 94.08 & 68.94 & 83.48 & 61.90 \\
		\rowcolor{gray!15} Text-IF$^{*}$ & 89.21 & 56.22 & 83.70 & 76.64 & 87.17 & 93.47 & 70.76 & 84.89 & 63.10 \\		
		ControlFusion & 88.52 & 49.54 & 84.34 & 76.57 & 86.27 & 94.16 & 71.00 & 84.20 & 61.67 \\
		\rowcolor{gray!15} ControlFusion$^{*}$ & 89.11 & 52.07 & 84.21 & 77.00 & 87.14 & 93.96 & 71.47 & 85.73 & 63.22 \\		
		URFusion      & 86.51 & 50.99 & 84.23 & 77.10 & 86.74 & 94.50 & 69.65 & 83.71 & 62.08 \\
		\rowcolor{gray!15} URFusion$^{*}$ & 88.11 & 53.09 & 83.54& 77.98 & 87.48 & 94.02 & 67.98 & 85.70 & 62.85 \\		
		TDFusion      & 88.01 & 53.27 & 84.36 & 76.25 & 86.66 & 94.04 & 69.23 & 83.94 & 62.14 \\
		\rowcolor{myLightBlue} TDFusion$^{*}$ & 89.58 & 54.78 & 83.86 & 78.00 & 87.12 & 94.20 & 68.95 & 85.64 & \underline{63.35} \\		
		Ours          & 87.85 & 48.26 & 85.01 & 77.39 & 87.10 & 94.54 & 69.91 & 83.28 & 62.27 \\
		\rowcolor{myLightRed} Ours$^{*}$  & 90.14 & 57.36 & 85.44 & 77.75 & 87.81 & 94.92 & 71.73 & 86.15 & \textbf{63.87}\\		
		\bottomrule
	\end{tabular}
\end{table}

\begin{figure}[t]
	\centering
	\includegraphics[width=1\columnwidth]{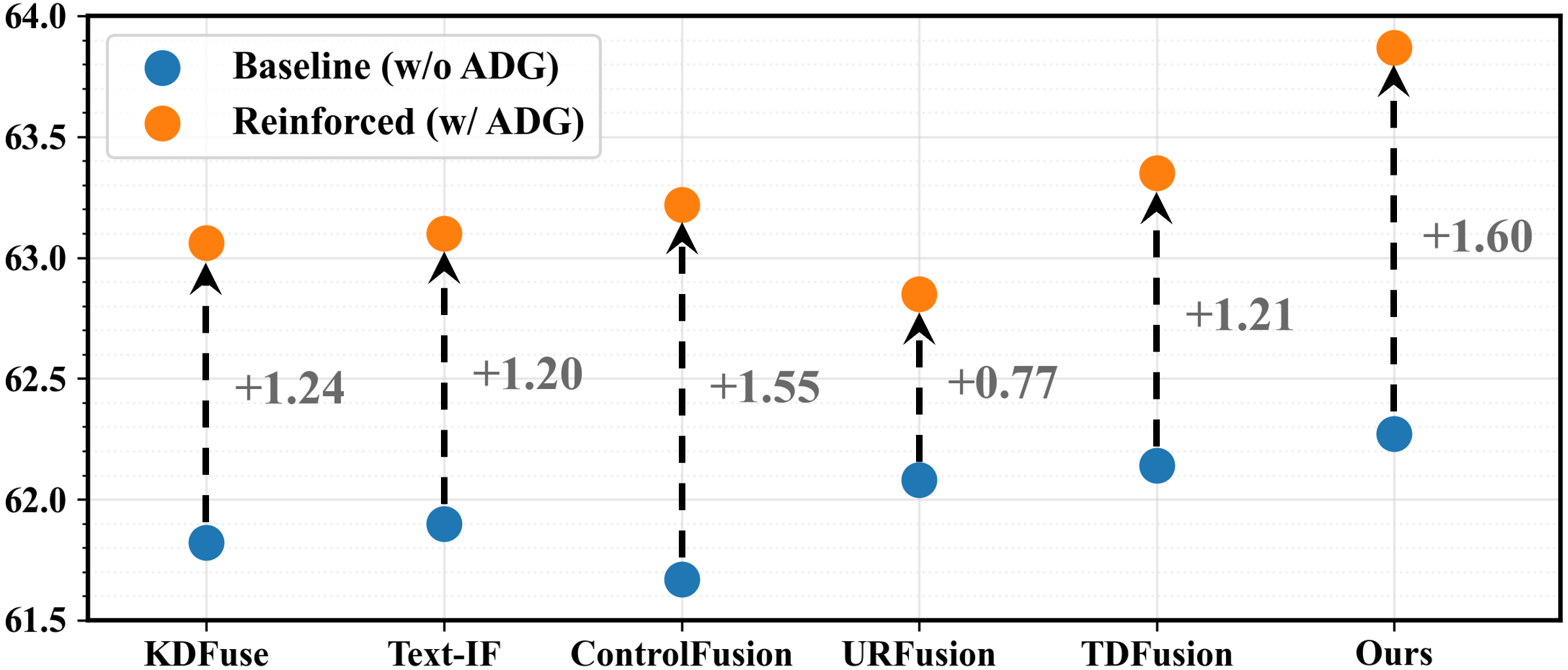} 
	\caption{Consistent semantic gain with ADG across multiple fusion methods.}
	\label{fig:13}
\end{figure}

\subsection{Generalization Validation of ADG}
To further validate the universal applicability and plug-and-play potential of the proposed ADG mechanism, we extend its deployment to five representative fusion methods. These methods are specifically selected because they demonstrate leading semantic performance as visual baselines. In this experiment, the fused images generated by various existing frameworks serve as the initial inputs, which are then refined through the ADG training paradigm to observe the potential for semantic reinforcement.
As illustrated in Table \ref{tab:Methods_Comparison} and Fig. \ref{fig:13}, the integration of the ADG mechanism yields consistent and significant performance boosts across all evaluated methods. It reveals that by repurposing the adversarial paradigm for explicit semantic reinforcement, the ADG mechanism effectively bridges the gap between low-level image fusion and high-level semantic requirements. These findings firmly establish ADG as a highly versatile and effective strategy, capable of empowering existing multimodal fusion systems to achieve superior semantic performance in complex scenarios.

\section{Conclusion and Discussion}

To bridge the long-standing gap between visual quality and semantic understanding in multimodal image fusion, this paper proposes a novel decoupled visual-semantic fusion paradigm, termed AD-SRF. We first construct a unified restoration-fusion baseline that not only effectively integrates cross-modal features but also exhibits robust degradation removal capabilities. This establishes a high-quality, information-rich visual foundation for both downstream tasks and human perception. To overcome the semantic bottlenecks inherent in traditional fusion models, we introduce an adversarial attack-defense game mechanism. By utilizing an attacker to probe semantic blind spots and driving a defender to perform targeted feature reinforcement, AD-SRF effectively restores and reinforces critical semantic information typically lost during the fusion process. Extensive experimental analysis demonstrates that AD-SRF achieves a win-win outcome in terms of both fused visual quality and semantic perception performance.

However, our method focuses on global semantic reinforcement, which may lack the flexibility to handle specific objects that require extra attention. Our future work will focus on making the reinforcement process more controllable. One promising solution is to use the vision-language models to integrate language instructions with the adversarial paradigm to achieve fine-grained, controllable semantic reinforcement.


\bibliographystyle{IEEEtran}
\bibliography{AD-SRF}

\vspace{-30pt}

\end{document}